\documentclass[journal]{IEEEtran}  
\ifCLASSINFOpdf
\else
\fi
\usepackage{amsmath,graphicx,cite}
\usepackage{subfig}
\usepackage{widetext}
\usepackage{amssymb}
\usepackage{rotating}
\usepackage{xcolor}

\begin{document}
%

\title{Robust Online Multi-target Visual Tracking using a HISP Filter with Discriminative Deep Appearance Learning}
%
%
%

\author{Nathanael~L.~Baisa*,~\IEEEmembership{Member,~IEEE,}
\thanks{*Nathanael L. Baisa is with the School of Computing and Communications, Lancaster University, Lancaster, LA1 4WA, UK. Part of this work was done when the author was with the School of Computer Science, University of Lincoln, Lincoln LN6 7TS, UK. (e-mail: nathanaellmss@gmail.com).}}
\maketitle

\begin{abstract}

We propose a novel online multi-target visual tracker based on the recently developed Hypothesized and Independent Stochastic Population (HISP) filter. The HISP filter combines advantages of traditional tracking approaches like MHT and point-process-based approaches like PHD filter, and it has linear complexity while maintaining track identities. We apply this filter for tracking multiple targets in video sequences acquired under varying environmental conditions and targets density using a tracking-by-detection approach. We also adopt deep CNN appearance representation by training a verification-identification network (VerIdNet) on large-scale person re-identification data sets. We construct an augmented likelihood in a principled manner using this deep CNN appearance features and spatio-temporal information. Furthermore, we solve the problem of two or more targets having identical label considering the weight propagated with each confirmed hypothesis. Extensive experiments on MOT16 and MOT17 benchmark data sets show that our tracker significantly outperforms several state-of-the-art trackers in terms of tracking accuracy.


\end{abstract}


%
\IEEEpeerreviewmaketitle

\section{Introduction} \label{sec:introduction}

Multi-target tracking is still an active area of research in computer vision with a broad range of applications including intelligent surveillance, autonomous driving, robot navigation and augmented reality. Its main goal is to estimate states from noisy observations and then associate them over-time to preserve the identity of targets. Tracking-by-detection is the most widely adopted paradigm for tracking multiple objects due to the emergence of more accurate object detection algorithms (using deep learning). In this approach, object detections are first obtained using object detectors in each frame and then the object trajectories are produced over-time using multi-target filters and/or data association. There are two methods to do this: online and offline tracking. In the online tracking approaches, the states of targets are estimated using Bayesian filtering~\cite{MatPoiCav16}\cite{SonJeo16}\cite{Nat19} at each time step using current observations, and miss-detections are handled by prediction using motion models to keep on tracking. However, offline (batch) approaches use both past and future observations to overcome miss-detections; mainly using global optimization-based data association~\cite{LeaCanSch16}\cite{MilRotSch14}\cite{PirRamFow11}. Even though offline methods perform better than the online methods, they are limited for time-critical real-time applications such as autonomous driving and robot navigation where it is crucial to deliver estimates as observations arrive.

There are unknown number of observations that can be generated as a result of applying an object detector (a sensor) to video frames which can be given as input to a multi-target tracker. In this detection process, there is uncertain information about clutter (false alarms), miss-detection and too close targets that are unresolved. This is basically related to the nature and origin of the observations caused by the object detector (sensor), and is thus referred to as measurement origin uncertainty~\cite{VoMalBarCorOsbMahVo15}. Not only this observation origin uncertainty, the multi-target tracker also has to deal with the objects' births (appearances), deaths (disappearances), and process and observation noises. Though including appearance information helps for resolving too spatially close objects, appearance variation due to illumination changes poses a challenge to the tracker. As noted in a survey of multi-target tracking algorithms~\cite{LuoZhaKim14}\cite{VoMalBarCorOsbMahVo15}, there are three well-known classical data association algorithms, Global Nearest Neighbor (GNN)~\cite{BarWilTia11}, Joint Probabilistic Data Association Filter (JPDAF)~\cite{BarWilTia11} and Multiple Hypothesis Tracking (MHT)~\cite{KimLiCip15,BarWilTia11}, that have been used for wide range of applications. Among these algorithms, the GNN, which can be computed using the Hungarian algorithm~\cite{FraJea71}, is generally the most computationally efficient algorithm with the lowest performance as it is sensitive to noise. The other two are computationally very expensive; the MHT is the most performing with the highest computational cost of all. Because of the complexity of these algorithms, another multi-target tracking paradigm which includes all sources of uncertainty in a unified framework has been proposed based on a random finite set (RFS) theory~\cite{Mah14}. From the RFS-based multi-target tracking algorithms, a probability hypothesis density (PHD) filter~\cite{Mah03} is the most widely used filter for video tracking in computer vision which has a linear complexity with the number of targets and measurements. However, the identity of targets is not included in this filter. Recently, a new filter, Hypothesized and Independent Stochastic Population (HISP) filter~\cite{HouCla18, DelHouFra17,DelHouFra19}, based on the stochastic populations has been proposed which includes all sources of uncertainty in a unified framework with a linear complexity while maintaining track identities.

More recently, astonishing results have been obtained on a range of recognition tasks using Convolutional Neural Network (CNN) features~\cite{KriHin12}\cite{KaiXiaSha15}. They have also shown better performance on object detection~\cite{ShaKaiRos15} and person re-identification~\cite{ZhengZY16} problems. There are also some tracking algorithms which have been proposed using deep learning~\cite{LeaCanSch16,AmiAleSil17} due to its powerful capturing capability of discriminative appearance features of a target of interest against background and other similar objects. However, the advantages of CNNs in stochastic population based filters, such as the HISP filter, have not been explored which works online and suitable for real-time applications.

In this work, we propose a novel online multi-object visual tracker which jointly addresses track management (target birth, death and labeling), miss-detection and false alarms (clutter) within a single Bayesian framework using a tracking-by-detection approach for time-critical real-time applications. Our proposed tracker is based on the recently developed HISP filter. We also learn discriminative deep appearance features of targets using a Verification-Identification CNN architecture (VerIdNet). We construct a single augmented likelihood using the learned deep CNN appearance representations and spatio-temporal (motion) information that can fit into the HISP filter for improving the performance of the tracker. We also overcome the problem of two or more objects having similar identity that occurs after the track extraction process in the HISP filter with the help of the weight of the confirmed hypothesis. To the best of our knowledge, this is the first work of the HISP Filter with deep representation learning.

The main contributions of this paper are as follows:
\begin{enumerate}
  \item We apply the HISP filter for tracking multiple objects in video frames captured under a range of environmental conditions and targets density.
  \item We learn discriminative deep appearance features of targets using the VerIdNet on large-scale person re-identification data sets.
  \item We construct an augmented likelihood using deeply learned CNN appearance features and motion information and then incorporate into the HISP filter.
  \item We solve the problem of two or more objects with similar label considering the weight propagated with each confirmed hypothesis.
  \item We make extensive evaluations on Multiple Object Tracking 2016 (MOT16) and MOT17 benchmark data sets using the public detections  given in the benchmark's test sets.
\end{enumerate}

A preliminary idea of this proposed approach was presented in~\cite{Nat18}. In this work, we give more explanations of our algorithm. Furthermore, we utilize deep CNN appearance features in addition to motion information to construct a single augmented likelihood that can fit into the HISP filter. We also make extensive evaluations on MOT17 besides MOT16 benchmark data set.

The rest of this paper is organized as follows. After the discussion of related work in section~\ref{sec:RelatedWork}, the HISP filter in visual tracking context is explained in detail including the modeling of the augmented likelihood in section~\ref{sec:HISP}. The discriminative deep appearance learning using VerIdNet is described in section~\ref{Sec:AppearanceLearning}, and section~\ref{Sec:VariableValues} provides some important variable values in the HISP filter implementation. The experimental results are analyzed and compared in section~\ref{Sec:ExperimentalResults} followed by the summary of the main conclusions along with suggestions for future work in section~\ref{Sec:Conclusion}.

\section{Related Work} \label{sec:RelatedWork}

There are many works on multi-object tracking in the literature that have been developed using the traditional data association-based Bayesian multi-target filters such as JPDAF~\cite{BarWilTia11} and MHT~\cite{KimLiCip15, BarWilTia11}. These methods extend single target tracking to multi-target tracking through data association. In these methods, finding the associations between objects and observations is the most challenging part due to two reasons. First, there is uncertainty caused by the data association which poses some wrong mappings of targets to measurements. Second, the algorithmic complexity increases exponentially with the number of objects and observations. Generally, these Bayesian data association-based multi-target tracking algorithms can be categorised into two~\cite{BarWilTia11}: target-oriented and measurement-oriented. In the target-oriented approach, the source of a measurement is assumed to be from either a known object or clutter, and this type of methods can not include the birth of targets within their framework; for instance JPDF. In the latter approach, the source of an observation is not only assumed to be from either a pre-existing target or clutter, but also from a newly appearing object. This type of algorithms include the appearance of new objects in their framework; for example MHT. Although the MHT includes the birth of new targets that enables it to track an unknown number of objects in the scene, it includes some heuristic and engineering solutions with a high computational demand and delayed association decisions which limit its applications for real-time applications.

Recently, a unified framework has been proposed by Mahler~\cite{Mah03} which extends single to multiple object tracking using RFS for the representation of multiple object states and measurements to address the problem of computational complexity. This approach includes all probabilistic uncertainties in the framework such as appearance and disappearance of objects, clutter and miss-detections. It simultaneously estimates the states and cardinality of unknown and time varying number of objects in the scene. The most adopted filter of this approach is the PHD filter~\cite{VoMa06} which propagates the first-order moment of the multi-object posterior referred to as the PHD (intensity)~\cite{VoMa06} rather than the full multi-object posterior. Another advantage of this approach is that it is flexible to be adapted for different applications. For instance, it has been applied to estimate the position of the detection proposal that has a maximum weight to track an object of interest not only in sparse but also in dense environments while suppressing the other detection proposals as false positives~\cite{BaiBhoWal18}\cite{NatDis18}. Furthermore, a novel N-type PHD filter ($N \geq 2$) has also been developed based on this standard PHD filter in~\cite{NatAnd19}\cite{BaiWal19} to track multiple objects of various types in the same scene. However, the downside of this approach is that the identity of an object is not included in the framework due to the indistinguishability assumption of the point process. To label each target, additional mechanism is mandatory either at the prediction stage~\cite{MatPoiCav16} or by post-processing the filter outputs~\cite{BaiWal19}. Recently, labeled RFS for multi-target tracking was introduced in~\cite{VoVoPhu14}\cite{VoVoHoa17}\cite{Kim17}, however, its computational complexity is high. In general, the RFS-based filters are susceptible to miss-detection even though they are robust to clutter.

More recently, the estimation framework for stochastic populations has been put forward which proposes a probabilistic representation of the population of interest through two levels of uncertainty: on the individual and population levels i.e. the targets of the population of interest are represented by tracks and the composition of the population of interest is represented by multi-target configurations. With the concept of partially-distinguishable populations, a stochastic populations-based filter has been introduced and is referred to as a Distinguishable and Independent Stochastic Populations (DISP) filter~\cite{DelHouCla16}. This filter can track an unknown and time-varying number of objects in the scene handling miss-detections, clutter, and appearance and disappearance of objects. Since this filter has a high computational cost, a low-complexity filter termed as a HISP filter~\cite{HouCla18} has been derived from the DISP filter under some intuitive approximations. Similar to the PHD filter, this HISP filter has a linear complexity with both the number of hypotheses and the number of measurements. However, it can maintain the identity of detected objects as opposed to the PHD filter.

There are also other non-Bayesian filtering-based multi-object tracking algorithms~\cite{BerFleTur11,WanTurFlu16,AndXinFra17,WanFanCha17} in the literature. A multi-object tracking algorithm has been developed in~\cite{BerFleTur11} based on the k-shortest paths algorithm for optimization which ensures a global optimum. A network-flow Mixed Integer Program (MIP) has been employed to develop a multi-target tracking algorithm in~\cite{WanTurFlu16} which allows to track different types of interacting objects. A method in~\cite{AndXinFra17} has used behavioral patterns to guide the tracking algorithm by imposing global consistency using non-Markovian approach. A batch-based Minimum-Cost Flows (MCF) has been proposed in~\cite{WanFanCha17} to develop a multi-target tracker. These all tracking methods~\cite{BerFleTur11,WanTurFlu16,AndXinFra17,WanFanCha17} are based on (global) optimization framework. Though such optimization-based approaches perform well, they are limited for time-critical real-time applications where it is important to deliver track estimates as measurements arrive.

Appearance information has been included in some works such as~\cite{KimLiCip15,KimLiReh18,Kim17,FuSngNaq18,Nat19}. Appearance models of targets are learned in an online fashion via a multi-output regularized least squares (MORLS) framework and are incorporated into the tree-based track-oriented MHT (TO-MHT) in~\cite{KimLiCip15}, and a bilinear long short-term memory (LSTM) has also been integrated into this MHT framework for gating in~\cite{KimLiReh18} by training it based on both motion and appearance. However, these trackers work offline and are computationally expensive. An approach similar to~\cite{KimLiCip15} has been employed in~\cite{Kim17} to learn the appearance models of targets and then to include this appearance information into a generalized labeled multi-Bernoulli (GLMB) filter. Though this GLMB-based tracker operates online, it is computationally expensive. Deep appearance features have also been incorporated into the PHD filter in~\cite{FuSngNaq18,Nat19}, however, the PHD filter is generally vulnerable to miss-detection though computationally efficient and robust to clutter. To date, no work has incorporated appearance information into the recently proposed stochastic populations-based filters and applied for visual tracking, as is the case in our work.

\section{The HISP Filter} \label{sec:HISP}

The HISP filter is derived from the high complexity DISP filter with a principled approximation for practical applications particularly for filtering a high number of objects with moderately ambiguous data association. It integrates the merits of engineering solutions such as MHT and point-process-based methods such as the PHD filter. By modeling all sources of uncertainties in a unified probabilistic framework, it solves the downsides of MHT including its high dependence on heuristics for the birth and death of objects and inflexibility while also propagating track identities through time similar to MHT. Furthermore, it has a linear complexity in the number of hypotheses and in the number of measurements, however, the MHT filter has an exponential complexity with time.

Let the time be indexed by the set $\mathbb{T}\doteq \mathbb{N}$. For any $t \in \mathbb{T}$, the object state space of interest and the observation space of interest are given by $\mathbf{X}^{\bullet}_t \subseteq \mathbb{R}^d$ and $\mathbf{Z}^{\bullet}_t \subseteq \mathbb{R}^{d{'}}$, respectively, where $d \geq d{'}$. They are augmented with the empty state $\psi_{es}$ and false-alarm generators $\psi_{fa}$ which describe the state of targets not in the area of interest and the targets not of direct interest while in the sensor's field of view but which may meddle with the measurement of the objects, respectively, and the empty observation $\phi$ which characterizes missed detections, to form the \textit{(full) target state space} $\mathbf{X}_t = \mathbf{X}^{\bullet}_t \bigcup \{\psi_{es}, \psi_{fa}\}$ and the \textit{(full) observation space} $\mathbf{Z}_t = \mathbf{Z}^{\bullet}_t \bigcup \{\phi\}$. The set of collected measurements is represented by $\bar{Z}_t = Z_t \bigcup \{\phi\}$; $Z_t$ for detected measurements.

At any time $t \in \mathbb{T}$, the HISP filter is based on the following modeling assumptions: 1) a target produces at most one measurement (if not, a miss detection occurs), 2) a measurement originates from at most one object (if not, a clutter occurs), 3) targets evolve independently of each other, and 4) observations resulting from target detections are produced independently from each other.

For tracking applications, objects are distinguished by considering their measurement histories. Let the space $\mathbb{O}_t$ be defined as the Cartesian product

\begin{equation}
\label{eq:eg}
\bar{\mathbb{O}}_t = \bar{Z}_0 \times ... \times \bar{Z}_t,
\end{equation}
\noindent
so that $\textbf{o}_t \in \mathbb{O}_t$ takes the form $\textbf{o}_t  = (\phi, ..., \phi, z_{t_+}, ..., z_{t_-}, \phi, ..., \phi)$ with $t_+$ and $t_-$ the time of birth and death of the considered track in the scene, and with $z_t \in \bar{Z}_t$ for any $t_+ \leq t \leq t_-$. The empty measurement path $(\phi, ..., \phi) \in \mathbb{O}_t$ is represented by $\boldsymbol\phi_t$ and $\textbf{o}_t$ is the measurement history (measurement path).

\begin{figure}
\centering
\begin{minipage}[c]{0.48\textwidth}
\centering
    \includegraphics[width=1.0\linewidth]{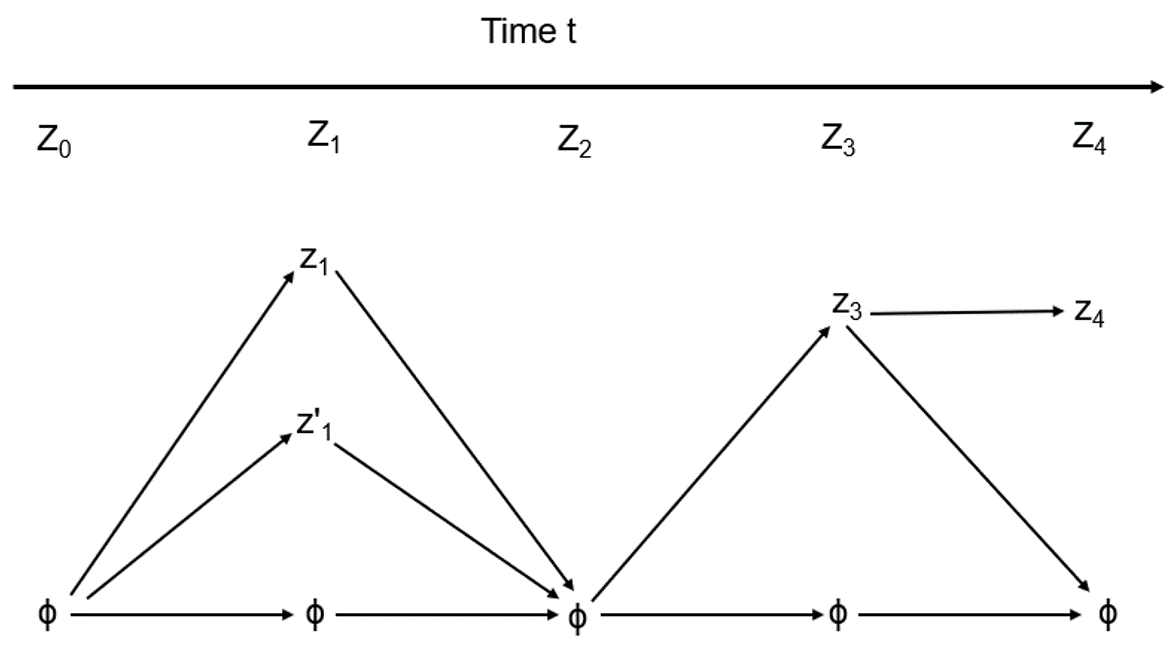}
    \caption{\small{The observation paths at time $t = 4$ of the HISP filter.}}
    \label{fig:HISPstructure}
\end{minipage}
\end{figure}

Each object is identified by some index $\textbf{\textit{i}}$ in a set $\mathbb{I}$. An observation characterizes an individual target. A track $\textbf{\textit{i}}$ associated to a measurement path with at least one detection (i.e. $\textbf{o}^{\textbf{\textit{i}}}_t \neq \boldsymbol\phi_t$) cannot have a multiplicity $n^{\textbf{\textit{i}}}$ greater than one as it cannot denote more than one object, Thus, the \textit{previously-detected} object represented by the track ${\textbf{\textit{i}}}$ is then \textit{distinguishable}. However, a track ${\textbf{\textit{i}}}$ associated to the empty measurement path $\textbf{o}^{\textbf{\textit{i}}}_t = \boldsymbol\phi_t$ denotes a sub-population of \textit{yet-to-be-detected} (undetected) objects that are \textit{indistinguishable} from one another, and may have a multiplicity $n^{\textbf{\textit{i}}}$ greater than one. The tracks cover all the possible combinations of non-empty observation paths representing the previously-detected targets, and one (or possibly several) track(s) representing sub-population(s) of yet-to-be-detected targets. For instance, the observation paths at time $t = 4$, given a sequence of collected observations, is shown in Fig.~\ref{fig:HISPstructure}. This diagrammatic illustration is somehow similar to a non-tree based track-oriented MHT~\cite{VoMalBarCorOsbMahVo15}.
\noindent
The observation paths (nine paths) in the diagram of Fig.~\ref{fig:HISPstructure} can be explicitly listed as:

\begin{gather}
(\phi, \phi, \phi, \phi, \phi) = \boldsymbol\phi      \\ \nonumber
(\phi, \phi, \phi, z_3, \phi)          \\  \nonumber
(\phi, \phi, \phi, z_3, z_4)         \\ \nonumber
(\phi, z{'}_1, \phi, \phi, \phi)  \\ \nonumber
(\phi, z{'}_1, \phi, z_3, \phi)   \\ \nonumber
(\phi, z{'}_1, \phi, z_3, z_4)   \\ \nonumber
(\phi, z_1, \phi, \phi, \phi)    \\ \nonumber
(\phi, z_1, \phi, z_3, \phi)     \\ \nonumber
(\phi, z_1, \phi, z_3, z_4)      \nonumber
\end{gather}

Each subset of pairwise compatible tracks $H \subseteq \mathbb{I}_t \setminus \{u\}$ which denotes the previously-detected objects is referred to as an hypothesis, and the set of all the hypotheses is denoted by $\textbf{H}_t$ whereas the undetected track $u$, with multiplicity $n^u \in \mathbb{N}$, represents sub-populations of $n^u$ yet-to-be-detected objects. Each element in the set $\mathbb{I}^u_t$ is represented by $\textbf{\textit{i}}^u_t$. The HISP filter maintain a single track $u \in \mathbb{I}$ for all the undetected targets. In the HISP filter, hypotheses are assumed to be independent of each other.

Accordingly, an object is indexed by a pair (t,\textbf{o}), i.e. \textbf{\textit{i}} = (t,\textbf{o}), where $t$ is the last epoch where the object was known to be in the scene, and the measurement path $\textbf{o}$ stores its detections across time. Thus, at any time $t \in \mathbb{T}$, the representation of objects after the prediction and after the update steps can be indexed by the sets $\mathbb{I}_{t|t-1} = \{(t,\textbf{o})| \textbf{o} \in \bar{\mathbb{O}}_{t-1}\}$ and $\mathbb{I}_t = \{(t,\textbf{o})| \textbf{o} \in \bar{\mathbb{O}}_t\}$, respectively. The other important notation is an indicator function, for instance, on the set A can be described as

\begin{equation}
    \mathbf{1}_A (x)=
\begin{cases}
    1,& \text{if } x\in A\\
    0,              & \text{otherwise}
\end{cases}
\label{eq:IndicatorFunction}
\end{equation}
\noindent

Using the previously mentioned concept and notations, the HISP filter can be described via a set of hypotheses. For example, after the measurement update step at time $t$ (see section~\ref{Sec:update}), it can be described by set of triples (multi-target configuration) of the form $\mathcal{P}_t = \{ p^{\textbf{\textit{i}}}_t, w^{\textbf{\textit{i}}}_t, n^{\textbf{\textit{i}}}_t\}_{{\textbf{\textit{i}}} \in \mathbb{I}_t}$, where $p^{\textbf{\textit{i}}}_t$ is the probability density corresponding to the index $\textbf{\textit{i}} \in \mathbb{I}_t$, $w^{\textbf{\textit{i}}}_t \in [0,1]$ is the probability of survival (weight) of the hypothesis, and $n^{\textbf{\textit{i}}}_t$ is the multiplicity of the hypothesis. Each hypothesis preserved by the HISP filter corresponds to a track (a confirmed hypothesis, see section~\ref{Sec:trackExtraction}) and is expressed by its own probability of survival.

The crucial steps of the HISP filter are concisely expressed as follows.

\subsection{Time Prediction}

\begin{figure}
\centering
\begin{minipage}[c]{0.48\textwidth}
\centering
    \includegraphics[width=1.0\linewidth]{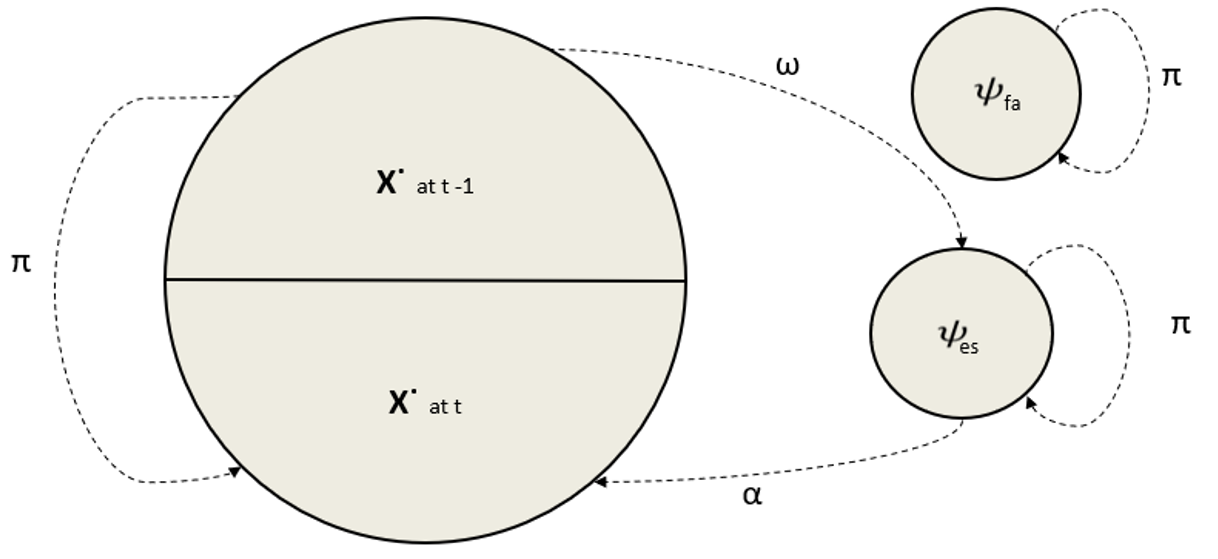}
    \caption{\small{The transition functions' relation between times t-1 and t for the subsets of $\mathbf{X}$.}}
    \label{fig:transitionFunctions}
\end{minipage}
\end{figure}

The three (sub-)transition functions used for modeling the dynamics (motion), birth and death of objects between times $t-1$ and $t$ are (also shown in Fig.~\ref{fig:transitionFunctions}):
\begin{enumerate}
  \item $q^{\alpha}_t$ models target's appearance (birth) i.e. transition from $\psi_{es}$ to $\mathbf{X}^{\bullet}$.
  \item $q^{\pi}_t$ models target's motion i.e. transitions from $\mathbf{X}^{\bullet}$ to $\mathbf{X}^{\bullet}$ or from $\psi_{es}$ or $\psi_{fa}$ to themselves.
  \item $q^{\omega}_t$ models target's disappearance (death) i.e. transition from $\mathbf{X}^{\bullet}$ to $\psi_{es}$.
\end{enumerate}


These functions are considered as sub-transitions since $q^{\iota}_t(.|x)$, $\iota \in \{\pi, \alpha, \omega\}$, is not a probability distribution although it is assumed to verify $\int q^{\iota}_t(x{'}|x)dx{'} \leq 1$. There is no transition between the subset $\mathbf{X}^{\bullet}_t \bigcup \{\psi_{es}\}$ describing the objects and the point $\psi_{fa}$ characterizing the clutter generators, thus, it holds that

\begin{equation}
q^{\iota}_t(x{'}|\psi_{fa}) =  q^{\iota}_t(\psi_{fa}|x) = 0,
\label{eq:transition_fa}
\end{equation}
\noindent for any $x, x{'} \neq \psi_{fa}$.

Accordingly, the motion of an object from time $t-1$ to time $t$ is modeled by a Markov transition $q^{\pi}_t$ satisfying for any $x{'} \in \mathbf{X}^{\bullet}_{t-1}$

\begin{equation}
q^{\pi}_t(\psi_{es}|\psi_{es}) = 1  ~~~~\text{and}~~~~  q^{\pi}_t(\psi_{es}|x{'}) = 0,
\label{eq:transition1}
\end{equation}
\noindent
The transition $q^{\pi}_t$ models propagation in the scene only (excluding object birth and death). The probability that an object at point $x$ at time $t-1$ does not disappear is given by the function $p^{\pi}_t (x) = \int q^{\pi}_t (x{'}|x)dx{'}$. The disappearance of an object between time $t-1$ and time $t$ is modeled separately by a transition $q^{\omega}_t$ satisfying for any $x{'} \in \mathbf{X}^{\bullet}_{t-1}$

\begin{equation}
q^{\omega}_t(x{'}|x)dx = 0  ~~~~\text{and}~~~~  q^{\omega}_t(x|\psi_{es})dx = 0,
\label{eq:transition1}
\end{equation}
\noindent
It is assumed that the transition $q^{\pi}_t$ and $q^{\omega}_t$ are complementary in the sense that $q^{\omega}_t (\psi_{es}|x) + p^{\pi}_t (x) = 1$, i.e. an object with state $x$ in $\mathbf{X}^{\bullet}$ can either propagate to $\mathbf{X}^{\bullet}$ with probability $p^{\pi}_t (x) = \int q^{\pi}_t (x{'}|x)dx{'}$ or disappear and move to $\psi_{es}$. Hence, the probability of existence of object with state $x$ is given by the scalar $p^{\pi}_t(x) = 1 - q^{\omega}_t (\psi_{es}|x)$. In addition, there are $n^{\alpha}_t$ objects potentially appearing at time $t$, modeled by a probability density $p^{\alpha}_t$ on $\mathbf{X}_t$ and by a scalar $w^{\alpha}_t$.

In the estimation framework for stochastic populations, the appearing objects and the yet-to-be detected (undetected) objects are mixed in a single sub-population. Using "$u$" in place of the indices $\textbf{\textit{i}}^u_{t-1}$ and $\textbf{\textit{i}}^u_{t|t-1}$ when there is no possible ambiguity, the newborn and the undetected objects are represented together after prediction by

\begin{subequations}
\begin{align}
        p^u_{t|t-1}(x)&=\frac{n^u_{t-1} \int q^{\pi}_t(x|x{'}) p^u_{t-1}(x{'})dx{'} + n^{\alpha}_t p^{\alpha}_t (x)}{n^u_{t-1} + n^{\alpha}_t},\\
        (w^u_{t|t-1},n^u_{t|t-1})&=\bigg(\frac{n^u_{t-1} w^u_{t-1} + n^{\alpha}_t w^{\alpha}_t}{n^u_{t-1} + n^{\alpha}_t}, n^u_{t-1} + n^{\alpha}_t \bigg),
\end{align}
\label{eq:newbornAndundetected}
\end{subequations}
\noindent
Where $p^{\alpha}_t (x) = q^{\alpha}_t (x|\psi_{es})$ is the distribution of appearing targets. The objects that have already been observed at least once in the past and which have prior indices in $\mathbb{I}_{t-1}$ of the form $\boldsymbol\kappa = (t-1, \textbf{o})$, with $\textbf{o}\neq \boldsymbol\phi_{t-1}$, can either be propagated (kernel $q^{\pi}_t$), and they are expressed after prediction by

\begin{subequations}
\begin{align}
        p^{\textbf{\textit{i}}}_{t|t-1}(x)&= \int q^{\pi}_t(x|x{'}) p^{\boldsymbol\kappa}_{t-1}(x{'})dx{'},\\
        w^{\textbf{\textit{i}}}_{t|t-1}&=w^{\boldsymbol\kappa}_{t-1}\int\int q^{\pi}_t(x|x{'}) p^{\boldsymbol\kappa}_{t-1}(x{'})dx{'}dx,\\
        n^{\textbf{\textit{i}}}_{t|t-1}&= 1,
\end{align}
\label{eq:surviving1}
\end{subequations}
\noindent or disappear (kernel $q^{\omega}_t$), and they are expressed after prediction by

\begin{subequations}
\begin{align}
        p^{\boldsymbol\kappa}_{t|t-1}(x)&= \int q^{\omega}_t(\psi_{es}|x{'}) p^{\boldsymbol\kappa}_{t-1}(x{'})dx{'},\\
        w^{\boldsymbol\kappa}_{t|t-1}&=w^{\boldsymbol\kappa}_{t-1}\int q^{\omega}_t(\psi_{es}|x{'}) p^{\boldsymbol\kappa}_{t-1}(x{'})dx{'},\\
        n^{\boldsymbol\kappa}_{t|t-1}&= 1,
\end{align}
\label{eq:surviving2}
\end{subequations}
\noindent with ${\textbf{\textit{i}}} =(t, \textbf{o}) \in \mathbb{I}_{t|t-1}$ (the object is still in the scene at epoch $t$) and $\boldsymbol\kappa =(t-1, \textbf{o})\in \mathbb{I}_{t-1}$ (the object has left the scene since last epoch $t-1$). Note that the index in $\mathbb{I}_{t|t-1}$ of the corresponding hypothesis of disappeared one remains equal to $\boldsymbol\kappa$.
The single-object laws of dead objects are not very informative, however, they are useful in practice as the scalar $\int q^{\omega}_t(\psi|x) p^{\boldsymbol\kappa}_{t-1}(x)dx$ provides the acceptability of the hypothesis that the object with index $\boldsymbol\kappa$ at time $t-1$ died between $t-1$ and $t$. The hypotheses corresponding to dead objects are not indexed in the set $\mathbb{I}_{t|t-1}$ as they are not considered for the following measurement update. However, it is necessary to store them since they will be helpful for track extraction (see section~\ref{Sec:trackExtraction}) although they are disregarded for the purpose of filtering.

The approximated multi-object configuration $\mathcal{P}_{t|t-1}$ after prediction from time $t-1$ to time $t$ is then provided by $\mathcal{P}_{t|t-1} = \{ p^{\textbf{\textit{i}}}_{t|t-1}, w^{\textbf{\textit{i}}}_{t|t-1}, n^{\textbf{\textit{i}}}_{t|t-1}\}_{{\textbf{\textit{i}}} \in \mathbb{I}_{t|t-1}}$. The time prediction step applies independently to each hypothesis as observed in the prediction equations~(\ref{eq:newbornAndundetected}),~(\ref{eq:surviving1}) and~(\ref{eq:surviving2}) because of the modeling assumption on the independence of the objects which makes it have a linear complexity with respect to the number of hypotheses.

\subsection{Measurement Update with Augmented Likelihood}\label{Sec:update}

The measurement process at time $t$ is modeled by a potential $\ell_t(z|.)$ on $\mathbf{X}_t$ defined for any $z \in \bar{Z}_t$ and satisfying $\ell_t(\phi|\psi_{es}) = 1$ as no measurement can be generated from objects that are not available in the scene. For any $x \in \mathbf{X}^{\bullet}_t$, the potential $\ell_t(z|.)$ can be provided by

\begin{equation}
\ell_t(z|x) = p_{d,t}(x) g_t(z|x), ~~ z \in Z_t  ~~~\text{and}~~~ \ell_t(\phi|x) = 1 - p_{d,t}(x),
\label{eq:update-ellz}
\end{equation}
\noindent
where $p_{d,t}$ is the probability of detection and the dimensionless potential $g_t(z|.)$ is the \textit{augmented likelihood} of association with observation $z$. In this case, let $z = (z_d, z_w)\in \bar{Z}_t$ be an ordered pair of a detection vector $z_d$ and an appearance feature vector $z_w$ which is extracted from the detected object region i.e. a region enclosed by a detection vector $z_d$ (object bounding box represented as a vector). Assuming a paired measurement $z \in \bar{Z}_t$ is an ordered pair conditionally independent observations, the augmented likelihood can be given by

\begin{equation}
g_t(z|x) = g_t(z_d|x) g_t(z_w|x),
\label{eq:AugmentedLikelihood}
\end{equation}
\noindent where the appearance feature likelihood function $g_t(z_w|x)$ will be discussed in Section~\ref{Sec:AppearanceLk}. The detection vector likelihood is provided in the two-dimensional, linear Gaussian case as

\begin{equation}
g_t(z_d|x) = \exp \bigg(-\frac{1}{2}(Hx - z_d)^T S^{-1} (Hx - z_d)\bigg) ,
\label{eq:update-lz}
\end{equation}
\noindent where $H$ is the measurement matrix and $S$ is the innovation covariance. With this approach, the probability for the observation $z$ to belong to the target with distribution $p$ is given by

\begin{equation}
\begin{array} {lll}
\int g_t(z|x)p(x)dx  =&  g_t(z_w|x) \sqrt{\frac{|R|}{|S|}} \exp \bigg(\\& -\frac{1}{2}(Hx - z_d)^T S^{-1} (Hx - z_d)\bigg) ,
\end{array}
\label{eq:likelihoodAss}
\end{equation}
\noindent which is dimensionless and takes values in the interval $[0, 1]$, and $|.|$ is the determinant. The $g_t(z_w|x)$ will be given in Section~\ref{Sec:AppearanceLk}.

To preserve a low computational cost for the HISP filter, all the terms in the measurement update can be computed with a linear complexity by making an assumption on the term

\begin{equation}
\breve{w}^{\boldsymbol\kappa,z}_t = w^{\boldsymbol\kappa}_{t|t-1} \int \ell_t (z|x) p^{\boldsymbol\kappa}_{t|t-1}(x) dx
\label{eq:update-Wbar}
\end{equation}
\noindent
which corresponds to the \textit{association weight} of the object with index $\boldsymbol\kappa \in \mathbb{I}_{t|t-1}$ with the measurement $z \in Z_t$.

For any $\boldsymbol\kappa = (t, \textbf{o}) \in \mathbb{I}_{t|t-1}$ and any $z \in \bar{Z}_t$, define $\textbf{\textit{i}}$ as the index $(t, \textbf{o} \times z)$, with $(\textbf{o} \times z)$ being the concatenation of $\textbf{o}$ and $z$, and define $p^{\textbf{\textit{i}}}_t$ as the probability density function on $\mathbf{X}_t$ expressed by

\begin{equation}
p^{\textbf{\textit{i}}}_t(x) = \frac{\ell_t(z|x) p^{\boldsymbol\kappa}_{t|t-1}(x)}{\int \ell_t(z|x{'}) p^{\boldsymbol\kappa}_{t|t-1}(x{'}) dx{'}}
\label{eq:updateP}
\end{equation}
\noindent
for any $x \in \mathbf{X}_t$, and let the weights be described equivalently by

\begin{equation}
w^{\textbf{\textit{i}}}_t = \frac{w^{\boldsymbol\kappa,z}_{e\text{x}} \breve{w}^{\boldsymbol\kappa,z}_t}{\sum_{z{'} \in \bar{Z}_t} w^{\boldsymbol\kappa,z{'}}_{e\text{x}} w^{\boldsymbol\kappa,z{'}}_t} ~~~ \text{or}~~~
w^{\textbf{\textit{i}}}_t = \frac{w^{\boldsymbol\kappa,z}_{e\text{x}} \breve{w}^{\boldsymbol\kappa,z}_t}{\sum_{\boldsymbol\kappa{'} \in \mathbb{I}_{t|t-1}} w^{\boldsymbol\kappa{'},z}_{e\text{x}} w^{\boldsymbol\kappa{'},z}_t}
\label{eq:updateW}
\end{equation}
\noindent
where the scalar $w^{\boldsymbol\kappa,z}_t = \breve{w}^{\boldsymbol\kappa,z}_t + \mathbf{1}_\phi (z) (1 - w^{\boldsymbol\kappa}_{t|t-1})$ is the probability mass allocated to the association between $\boldsymbol\kappa$ and $z$ including the possibility that the object does not actually exist in the case of miss-detection. The probability that a clutter will be generated for $z \in \bar{Z}_t$ is represented by $v_t(z)$. The posterior probability for a measurement $z \in Z_t$ to be a clutter is also obtained via Eq.~(\ref{eq:updateW}) when $\boldsymbol\kappa = z$, by setting $w^{z,z}_t = \breve{w}^{z,z}_t = v_t(z)$, $w^{z,\phi}_t = 1 - v_t(z)$, and $w^{z,z{'}}_t = 0$ if $z \neq z{'}$. For any $z \in \bar{Z}_t$ and any $\boldsymbol\kappa \in \mathbb{I}_{t|t-1}$ or $\boldsymbol\kappa = z$, the \textit{external weight} $w^{\boldsymbol\kappa,z}_{e\text{x}}$ (scalar) is the weight corresponding to the association of the measurements in $Z_t \setminus \{z\}$ with false positives, any of the remaining undetected individuals, or any remaining hypotheses in $\mathbb{I}_{t|t-1}\setminus \{\boldsymbol\kappa\}$ i.e. $w^{\boldsymbol\kappa,z}_{e\text{x}} \in [0,1]$, the multi-target marginal likelihood, can be seen as the assessment of the compatibility between the predicted laws ($\{ p^{\textbf{\textit{i}}}_{t|t-1}\}_{{\textbf{\textit{i}}} \in \mathbb{I}_{t|t-1}}$) and the collection of measurements at the current time excluding the object with index $\boldsymbol\kappa$ and the measurement $z$. This scalar can be characterized as

\begin{equation}
w^{\boldsymbol\kappa, z}_{e\text{x}} = C{'}_t (\boldsymbol\kappa, z) \prod_{\boldsymbol\kappa{'} \in \mathbb{I}_{t|t-1} \setminus \{\boldsymbol\kappa\}} \bigg[w^{\boldsymbol\kappa{'}, \phi}_t + \sum_{z{'} \in Z_t \setminus \{z\}} \frac{w^{\boldsymbol\kappa{'}, z{'}}_t}{C_t(z{'})} \bigg]
\label{eq:updateW-ex}
\end{equation}
\noindent
where $C_t(z) = w^{u,z}_t/w^{u,\phi}_t + v_t(z)/(1-v_t(z))$ and where

\begin{equation}
\begin{array} {lll}
C{'}_t (\boldsymbol\kappa, z) =& [w^{u,\phi}_t]^{n^u_{t|t-1} - \mathbf{1}_u(\boldsymbol\kappa)} \\& \bigg[\prod_{z{'}\in Z_t\setminus Z{'}} (1-v_t({z{'}}) \bigg] \bigg[\prod_{z{'} \in Z_t\setminus \{z\}} C_t(z{'}) \bigg]
\end{array}
\label{eq:updateW-C}
\end{equation}
\noindent
with $Z{'} = \emptyset$ when $\boldsymbol\kappa \in \mathbb{I}_{t|t-1}$ and $Z{'} = \{z\}$ when $\boldsymbol\kappa$ corresponds to a clutter $(\boldsymbol\kappa = z)$. The hypotheses corresponding to false positives are not indexed in the set $\mathbb{I}_t$ since they are not considered for the next time step. However, it is necessary to store them since they will be helpful for track extraction (see section~\ref{Sec:trackExtraction}) although they are disregarded for the purpose of filtering.

The approximated multi-object configuration $\mathcal{P}_t$ after the observation update at time $t$ is then provided by $\mathcal{P}_t = \{  p^{\textbf{\textit{i}}}_t, w^{\textbf{\textit{i}}}_t, n^{\textbf{\textit{i}}}_t\}_{{\textbf{\textit{i}}} \in \mathbb{I}_t}$ where $n^{\textbf{\textit{i}}}_t = n^u_{t|t-1}$ if $\textbf{\textit{i}} = u$ and $n^{\textbf{\textit{i}}}_t = 1$ otherwise. It can also be stated that the collection of marginalized single-object laws $\{p^{\textbf{\textit{i}}}_t\}_{{\textbf{\textit{i}}} \in \mathbb{I}_t}$ can be treated as single-object filters in interaction, where the weights of the filters are $\{w^{\textbf{\textit{i}}}_t\}_{{\textbf{\textit{i}}} \in \mathbb{I}_t}$. The formulation of the posterior weights Eqs.~(\ref{eq:updateW}), (\ref{eq:updateW-ex}) of the hypotheses is obtained using two assumptions. The first one is for any $\boldsymbol\kappa, \boldsymbol\kappa{'} \in \mathbb{I}_{t|t-1}$ such that $\boldsymbol\kappa \neq \boldsymbol\kappa{'}$ and any $z \in Z_t$, it holds that $\breve{w}^{\boldsymbol\kappa,z}_t \breve{w}^{\boldsymbol\kappa{'},z}_t \approx 0$. This indicates that the data association is moderately ambiguous i.e. it is related to sparsity of the scenario; the HISP filter is based on the sparsity assumption. The second assumption is that hypotheses are independent of each other. Especially, the computation of the weight $w^{\boldsymbol\kappa,z}_{e\text{x}}$ does not involve combinatorial operations on the subsets of measurements and/or hypotheses which makes the measurement update step have a linear complexity with respect to the number of hypotheses and the number of measurements i.e. $\mathcal{O}(|\mathbb{I}_{t|t-1}||Z_t|)$. Thus, a complexity of the HISP filter complete recursion is $\mathcal{O}(|\mathbb{I}_{t|t-1}||Z_t|)$ because the computation of the terms $w^{\boldsymbol\kappa, z}_{e\text{x}}$ is the only part of it with a higher complexity.

\subsection{Pruning and Merging}

As mentioned earlier, the HISP filter has a linear complexity with regards to the number of hypotheses and the number of observations. Still giving a restriction on the number of propagated hypotheses without a reasonable information loss is necessary without affecting a track extraction. Thus, pruning and merging of these hypotheses are important which are, in this case, applied to the output of the HISP filter at time $t$ with a multi-object configuration $\mathcal{P}_t = \{  p^{\textbf{\textit{i}}}_t, w^{\textbf{\textit{i}}}_t, n^{\textbf{\textit{i}}}_t\}_{{\textbf{\textit{i}}} \in \mathbb{I}_t}$, and are described as follows:

\begin{enumerate}
  \item Some hypotheses are more important than the others determined by their weight. Thus, a hypothesis $\textbf{\textit{i}} \in \mathbb{I}_t$ may have insignificant weight $w^{\textbf{\textit{i}}}_t$ which can be pruned by keeping the subset of hypotheses that have a weight greater than a threshold of $\tau_p$.
  \item Probability densities $p^{\textbf{\textit{i}}}_t$, $\textbf{\textit{i}} \in I$ of some hypotheses $I \subseteq \mathbb{I}_t$ are very near to each other which may represent very alike information. We merge the probability densities that have less than a threshold of $\tau_m$ Mahalanobis distance computed between their distributions.
  \item Some hypotheses $I \subseteq \mathbb{I}_t$ may have the same observation path over the extraction window $T$. It is possible to merge such hypotheses into a single hypothesis with $w = \sum_{\textbf{\textit{i}} \in I} w^{\textbf{\textit{i}}}_t $ if $w \leq 1$ as hypotheses cannot have a weight strictly greater than 1. This is with the assumption that they may characterize the same potential object.
\end{enumerate}
After the pruning and merging steps, the multi-object configuration will be $\tilde{\mathcal{P}}_t = \{ \tilde{p}^{\textbf{\textit{i}}}_t, \tilde{w}^{\textbf{\textit{i}}}_t, \tilde{n}^{\textbf{\textit{i}}}_t\}_{{\textbf{\textit{i}}} \in \tilde{\mathbb{I}}_t}$, and is utilized in the next time step.

\subsection{Track Extraction}\label{Sec:trackExtraction}

The subset of hypotheses propagated by the HISP filter which most likely represents objects in the scene is referred to as tracks. We extract these tracks from the approximated multi-object configuration without modifying the set of hypotheses. Thus, the track extraction method is used only for output purpose without causing any effect on the filtering process. Accordingly, we select the subset of hypotheses with the highest possible weights and whose observation paths are in agreement with the measurements collected during a sliding time window $T$. Since it is crucial to be aware of all observations gathered in a given time window for the track extraction purpose, we need to store the posterior probabilities of each observation to be a clutter as hypotheses. Along with these, we also need to store the hypotheses corresponding to disappeared objects in this time window. Provided the temporary set of hypotheses $\hat{\mathbb{I}}_t$ obtained from the above modifications, we can solve the track extraction using the following optimization problem

\begin{equation}
\underset{I \subseteq \hat{\mathbb{I}}_{t}}{\operatorname{argmax}} \prod_{\textbf{\textit{i}} \in I} \tilde{w}^{\textbf{\textit{i}}}_t
\label{eq:trackExtraction}
\end{equation}
\noindent
subject to 1) all the measurements must be contained in the union of all measurement paths in the time window $T \subseteq \mathbb{T} \cap [0,t]$, and 2) the observation paths in $I$ must be pairwise compatible i.e. each measurement should not be used more than once. The solution to the problem in Eq.~(\ref{eq:trackExtraction}) is the same as the one for

\begin{equation}
\underset{I \subseteq \hat{\mathbb{I}}_{t}}{\operatorname{argmax}} \sum_{\textbf{\textit{i}} \in I} \log \tilde{w}^{\textbf{\textit{i}}}_t
\label{eq:trackExtraction2}
\end{equation}
\noindent
with similar constraints as all $\tilde{w}^{\textbf{\textit{i}}}_t$ are strictly positive. Taking in this way is helpful as it allows us to solve it using integer programming such as the GNU Linear Programming Kit (GLPK). This optimization problem is used to select the tracks which have been observed at least once in the given sliding window. Although there are many track extraction methods available, the used one is one of the simplest and efficient approach which uses the structure of the HISP filter, for instance, rather than selecting the hypotheses based on their weight on individual basis.

In the context of video tracking, there is a possibility that many closely located objects can be treated as a single object by the object detector with one bounding box response to all of them because of their extended nature. This usually happens when density of the objects is very high. These objects can be detected by their corresponding bounding boxes when they begin to separate. This is similar to the situation where some objects spawn from the original one. Thus, when we extract the tracks using the aforementioned method, spawning objects may be given identical label to the original object. This creates a problem of differentiating them since they share similar label. In this work, we solve this problem by utilizing the weight propagated with each confirmed hypothesis, also referred to as track, for discriminating them. In this case, we assume that the original object, after the track extraction procedure, has a maximum weight, and we also treat the spawned objects as new births of tracks. Accordingly, we assign new label(s) to the spawned object(s) if two or more tracks are retrieved from the track extraction process with similar label by assuming that the original object has a maximum weight and thus deserves to keep the original label. Although, this approach overcomes the problem of this similar track labeling issue, it is rarely susceptible to identity switches. This is due to the violation of our assumption i.e. the spawned object(s) can have weight(s) greater than the original object. Note that this process does not have any effect on the filtering process as it is barely for output purpose.

\section{Discriminative Deep Appearance Learning} \label{Sec:AppearanceLearning}

Visual appearance information is very crucial for tracking multiple targets as it provides rich information to disambiguate nearby targets where only spatio-temporal information can not handle. In this work, we learn deep appearance features of targets using deep CNN and incorporate into the likelihood function of the HISP filter. We adopt ResNet50~\cite{KaiXiaSha15} as the backbone network of our Verification-Identification architecture to learn appearance information of targets from a large-scale training data set.

\subsection{Appearance Feature Likelihood} \label{Sec:AppearanceLk}

The appearance feature likelihood function $g_t(z_w|x)$ is given by

\begin{equation}
g_t(z_w|x) =  \frac{\exp(C^w(\mathbf{z}^i_w, \mathbf{z}^j_w))}{\exp(C^w(\mathbf{z}^i_w, \mathbf{z}^j_w)) + \exp(-C^w(\mathbf{z}^i_w, \mathbf{z}^j_w))},
\label{eq:AppearanceLikelihood}
\end{equation}
\noindent where $C^w(\mathbf{z}^i_w, \mathbf{z}^j_w)$ is the cosine distance between appearance feature vectors $\mathbf{z}^i_w$ and $\mathbf{z}^j_w$ which are extracted from the detected object regions determined by the detection vectors $z^i_d$ and $z^j_d$, respectively, using the learned deep appearance model in section~\ref{Sec:VerIdNet}. As can be seen from Eq.~(\ref{eq:likelihoodAss}), $z^i_d$ corresponds to the predicted measurement vector $Hx$ and $z^j_d$ corresponds to the current detection vector $z_d$. The cosine distance $C^w(\mathbf{z}^i_w, \mathbf{z}^j_w)$ between appearance feature vectors $\mathbf{z}^i_w$ and $\mathbf{z}^j_w$ is given using their dot product and magnitude (norm) as

\begin{equation}
C^w(\mathbf{z}^i_w, \mathbf{z}^j_w) =  \frac{\mathbf{z}^i_w \cdot \mathbf{z}^j_w}{\| \mathbf{z}^i_w \| \|\mathbf{z}^j_w\|},
\label{eq:CosineD}
\end{equation}
\noindent Since this appearance feature likelihood is combined with the detection vector likelihood in Eq.~(\ref{eq:AugmentedLikelihood}), it is utilized for both data association and track updates.

\subsection{Verification-Identification Network-Based Appearance Learning} \label{Sec:VerIdNet}

\begin{figure*}[htbp] 
\begin{center}
  \includegraphics[width=1.0\linewidth]{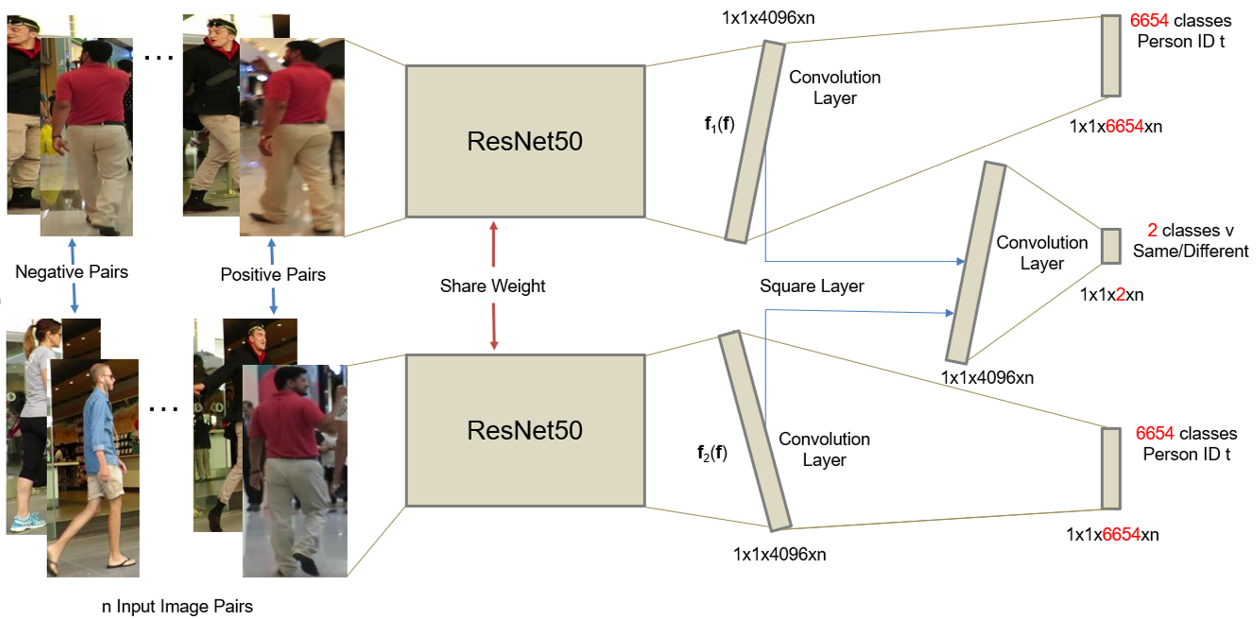} \\
\end{center}
   \caption{Illustration of VerIdNet for discriminative deep appearance learning using two ResNet50 models.}
\label{fig:VerIdNet}
\end{figure*}

We use ResNet50 in our Verification-Identification Network (VerIdNet) which is illustrated in Fig.~\ref{fig:VerIdNet}. This network is basically a Siamese network that integrates the verification and identification losses i.e. it optimizes three objectives: two identification losses and one verification loss. Though this type of architecture has been used in face recognition community~\cite{SunWanTan14}, ours is more similar to~\cite{ZhengZY16}. However, unlike the one in~\cite{ZhengZY16} which learns a re-identification appearance model on only one data set (Market1501~\cite{ZheSheTia15} with 751 identities), we trained this network on agglomeration of data sets (which has a total of 6,654 identities) with longer epochs to learn more robust discriminative appearance features of targets.

Basically, verification and identification models are the two well known types of CNN structures. The verification model performs a binary classification which determines whether a pair of images belong to the same person or not i.e. it is a similarity metric. However, it does not consider the relationship between the image pairs and the other images in the data set (all annotated information is not considered). On other hand, identification model takes full advantages of all annotated labels and treats the task as a multi-class recognition problem to learn a discriminative CNN embedding\footnote{An embedding is low-dimensional space into which a high-dimensional data can be mapped to i.e. semantically similar objects are grouped together in the embedding space.}, however, it does not consider the similarity between image pairs. The VerIdNet uses the merits of the two models for learning more robust discriminative appearance features of targets.

\noindent

As shown in Fig.~\ref{fig:VerIdNet}, the VerIdNet uses two ResNet50 models pre-trained on ImageNet~\cite{ILSVRC15} by removing the final fully-connected (FC) layer, and with additional three Convolutional Layers, Square Layer and three losses. The Square Layer compares the high-level 4096-dimensional embeddings by subtracting and squaring element-wisely. Moreover, softmax and cross-entropy loss are used in this VerIdNet. Thus, given a pair of images, the VerIdNet simultaneously predicts the IDs of the two images and their similarity score.

\textbf{Data preparation: }
We collect the training data from multiple sources. From all the training data set of MOT15 (TUD-Stadmitte, TUD-Campus, PETS09-S2L1, ETH-Bahnhof and ETH-Sunnyday) and MOT16/17 (5 sequences), we produce about 521 person identities. MOT16 and MOT17 have the same training data set though MOT17 is claimed to have more accurate ground truth and is used in our experiment. From TownCentre data set~\cite{BenRei11}, we also produce about 213 identities. In addition to these tracking data sets, we also use publicly available person re-identification data sets such as Market1501 data set~\cite{ZheSheTia15} (736 identities from 751 as we restrict the number of images per identity to at least 4), DukeMTMC data set~\cite{RisSolZou16} (702 identities), CUHK03 data set~\cite{LiZhaXia14} (1367 identities), LPW data set~\cite{sonLenLiu17} (1974 identities), and MSMT data set~\cite{LonShiWen17} (1141 identities). Including the tracking data sets into these public re-identification data sets for training our network allows the network to learn the inter-frame variations which is useful for improving the tracking performance. From all these data sets, we produce about 6,654 identities for training our network. 10\% of this training set (of each person identity if the number of images for that identity is greater than 9) is used for validation; no validation for small class.

We resize all the training images to $256\times256$ and then subtract the mean image from all the images; the mean image is computed from all the training images. During training, we randomly crop all the images to $224\times224$ and then mirror horizontally. We use a random order of images by reshuffling the data set. The positive pairs are generated by randomly sampling image patches (cropped images) from the same person identities whereas the negative pairs are the image patches corresponding to different classes/identities. We set the ratio between the positive and negative pairs to 1:1 initially and then multiply it by a factor of 1.01 every epoch until it arrives 1:4, similar to the work in~\cite{ZhengZY16}. This is due to the limited number of positive pairs that may cause the network to over-fit.

\textbf{Training: }
We train the VerIdNet using softmax (to normalize the output), cross-entropy loss and mini-batch Stochastic Gradient Descent (SGD). After computing all the gradients produced by every objective (there are three objectives in the VerIdNet), we add the weighted gradients together (1.0 for verification loss and 0.5 for the two identification losses) to update the network. In addition, we augment the training samples by random flipping horizontally as well as randomly shifting the cropping locations by no more than $\pm 0.2$ of detection box of width and height for $x$ and $y$ dimensions, respectively, to increase more variation and thus reduce possible over-fitting. We use a dropout of 0.75 (setting the output of each hidden neuron to zero with probability 0.75) before the final convolution layer for reducing a possible over-fitting.

\textbf{Testing: }
Two video sequences of MOT16/17 training data set (02 and 09) are used for testing of the VerIdNet. For this testing set, we produce about 66 person identities. As depicted in Fig.~\ref{fig:VerIdNet}, the two ResNet50 models share weights, therefore, we extract features by activating only one fine-tuned ResNet50 model. Thus, we get a 4,096-dimensional descriptor \textbf{f} given a $224 \times 224$ input image by feeding it forward to the ResNet50 in our network. Then, a cosine distance is computed between the extracted appearance feature vectors as in section~\ref{Sec:AppearanceLk}. For evaluating the trained VerIdNet, we randomly sample about 800 positive pairs (the same identities) and 3200 negative pairs (different identities) from ground truth of MOT16/17-02 and MOT16/17-09 training data set (used as a testing set for our network). We use this larger ratio of negative pairs to mimic the positive/negative distribution during tracking. We use verification accuracy as an evaluation metric. Given a pair of images, the cosine distance (using Eq.~(\ref{eq:CosineD})) between their extracted deep appearance feature vectors obtained using the learned model is computed. If the computed cosine-distance of positive pairs is greater than or equal to 0.75, they are assumed as correctly classified pairs. Similarly, if the computed cosine-distance of negative pairs is less than 0.75, they are assumed as correctly classified pairs. Accordingly, the VerIdNet trained on large-scale data sets (6,654 identities) gives about 98\% accuracy while the work in~\cite{ZhengZY16} trained on only Market150 data set (751 identities) gives about 86\% accuracy; our learned model gives a significant improvement.

\section{Variable Values in the HISP Filter Implementation} \label{Sec:VariableValues}

It is possible to implement the HISP filter using any Bayesian filtering technique for each hypothesis such as sequential Monte Carlo (SMC)~\cite{JerDanPie15} or Kalman filter (KF). Our work is based on KF implementation of the HISP filter (KF-HISP). In this linear Gaussian model assumption, a probability density $p^{\textbf{\textit{i}}}_t$ is described by a multivariate normal distribution $\mathcal{N}(m^{\textbf{\textit{i}}}_t, P^{\textbf{\textit{i}}}_t)$ where $m^{\textbf{\textit{i}}}_t$ is the mean and $P^{\textbf{\textit{i}}}_t$ is the covariance for $\textbf{\textit{i}} \in \mathbb{I}_t$.

We include centroid positions, velocities, width and height of the bounding boxes of detections in our state vector $x_t = [p_{cx,xt}, p_{cy,xt}, \dot{p}_{x,xt}, \dot{p}_{y,xt}, w_{xt}, h_{xt}]^T$ which is to be estimated using the filter. Similarly, the measurement is the centroid positions, width and height of the bounding boxes of detections which is assumed to be the noisy version of the state vector. Thus, the observation vector is $z_t = [p_{cx,zt}, p_{cy,zt}, w_{zt}, h_{zt}]^T$.

An object state evolves from time (frame) $t-1$  to $t$ with the help of the Markov transition kernel $q^{\pi}_t$. In this case, the state transition matrix $F$ and process noise covariance $Q$ take into consideration the bounding box width and height at the provided scale as follows,

\[ F_{t-1} = \left[ \begin{array}{ccc}
           I_2 & \Delta I_2 & 0_2 \\
           0_2 & I_2 & 0_2 \\
           0_2 & 0_2 & I_2
           \end{array} \right], \]
													
\begin{equation} Q_{t-1} = \sigma_v^2 \left [ \begin{array}{ccc}
    \frac{\Delta^4}{4}I_2 & \frac{\Delta^3}{2}I_2  & 0_2\\
    \frac{\Delta^3}{2}I_2 & \Delta^2 I_2 & 0_2 \\
    0_2 & 0_2 & I_2
    \end{array} \right],
\label{eqn:PHDstateTransitionMatrixVideo}
\end{equation}
\noindent
where $I_n$ and $0_n$ represent, respectively, the \textit{n} x \textit{n} identity and zero matrices. $\Delta = 1$ second is the sampling period between consecutive frames. $\sigma_v = 5$ pixels$/s^2$ is the standard deviation of the process noise. We assume the disappearance kernel $q^{\omega}_t$ to be constant which satisfies, for any $x \in \mathbf{X}^{\bullet}_t$, $q^{\omega}_t(\psi_{es}|x) = 10^{-2}$. This means the probability of existence $p^{\pi}_t$ of the objects is 0.99. It is crucial to use appropriate value of $p^{\pi}$ as the HISP filter is sensitive to it. For instance, $p^{\pi} = 1$ implies that if an hypothesis is available somehow certainly, it will be output at all consecutive time steps. Otherwise, hypotheses halt to be treated as tracks as soon as a miss-detection occurs if $p^{\pi}_t \leq p_d$. Therefore, to handle some miss-detections, it is necessary to assign the value of $p^{\pi}$ greater than the value of the probability of detection $p_d$.

Likewise, the observation follows the observation model, Eq.~(\ref{eq:update-ellz}). The measurement matrix $H_t$ and the observation noise covariance $R_t$ take into consideration the bounding box width and height as follows,

\[H_t = \left[ \begin{array}{ccc}
           I_2 & 0_2 & 0_2 \\
           0_2 & 0_2 & I_2
           \end{array} \right], \]

\begin{equation} R_t = \sigma_r^2 \left [ \begin{array}{cc}
     I_2 & 0_2 \\
     0_2 & I_2
    \end{array} \right],
\label{eqn:eqn:PHDobservationMatrixVideo}
\end{equation}
\noindent
where $\sigma_r = 6$ pixels is the observation standard deviation. We set the value of the probability of detection $p_d = 0.90$ with the assumption of a constant value across the state space and through time. We also assume the false positives (alarms) to be independently and identically distributed (i.i.d), and the number of false alarms per frame is Poisson-distributed with mean 10 (clutter rate of $v_t(z) \approx 4.8 \times 10^{-6}$; division of the mean 10 by frame resolution).

We set the mean number of newly appearing objects per frame $n^{\alpha}_t$ to $0.1$. The probability $w^{\alpha}_t$ which any potential measurement representing an object birth is obtained by dividing $n^{\alpha}_t$ uniformly across frame resolution. The distribution $p^{\alpha}_t$ is uninformative because nothing is known about the targets birth before the first measurement. Using a predetermined initial covariance provided in Eq.~(\ref{eqn:PHDbirthCovariance}) for objects birth and the current measurement and zero initial velocity utilized as a mean of the Gaussian distribution, the distribution after the measurement is set.

\begin{equation}
 P^{\alpha}_t  = diag([100, 100, 25, 25, 20, 20]).
\label{eqn:PHDbirthCovariance}
\end{equation}
\noindent

For reducing the computational demand of the HISP filter, we set the pruning threshold $\tau_p$ to $10^{-3}$ and the merging threshold $\tau_m$ to 4 pixels. These threshold values are utilized on the collection of individual posterior laws (probability densities). We also set the sliding time window $T$ to 5. The maximum number of hypotheses is set to $10^7$.

\section{Experimental Results} \label{Sec:ExperimentalResults}

\textbf{Tracking Datasets: } We evaluate our proposed online tracker by conducting extensive experiments on the MOT16 and MOT17 benchmark data sets~\cite{MOT16}. These data sets are filmed on unconstrained environments using both static and moving cameras, and are widely used data sets for multiple people tracking. They consist of 7 training sequences for which ground-truth information is available and 7 testing sequences for making a fair comparison of multi-object tracking algorithms. Thus, we use the publicly released detections provided on the MOT Challenge website which were produced using DPM~\cite{FelGirMcA10}, FRCNN~\cite{ShaKaiRos15} and SDP~\cite{YanChoLin16} object detectors. We use these detections with a non-maximum suppression (NMS) of 0.3 for DPM detector (for both MOT16 and MOT17) and 0.5 for FRCNN and SDP detectors (for MOT17).

\textbf{Evaluation Metrics: } We use several metrics to evaluate the performance of our online tracker including the CLEAR metrics~\cite{BerSti08}, the identity preservation measure~\cite{RisSolZou16} and the set of track quality measures~\cite{LiHua09} which are described as follows:

\begin{itemize}
  \item Multiple Object Tracking Accuracy (MOTA): Overall tracking accuracy which combines false positives, false negatives and identity switches to provide a measure of the performance of a tracker at both detecting targets and preserving their identities.
  \item Multiple Object Tracking Precision (MOTP): Overall tracking precision that shows a tracker's capability in estimating accurate positions of objects determined by a bounding box overlap between tracked location and ground-truth.
  \item Identification F1 (IDF1) score: The quantitative measure obtained by dividing the number of correctly distinguished detections by the mean of the number of ground truth and detections.
  \item Mostly Tracked targets (MT): Percentage of mostly tracked targets (an object is tracked for not less than 80\% of its life-span regardless of maintaining its identity) to the total number of ground-truth trajectories.
  \item Mostly Lost targets (ML): Percentage of mostly lost targets (an object is tracked for less than 20\% of its life-span) to the total number of ground-truth trajectories.
  \item False Positives (FP): Number of false detections.
  \item False Negatives (FN): Number of miss-detections.
  \item Identity Switches (IDSw): Number of times the given label of a ground truth track alters.
  \item Fragmented trajectories (Frag): Number of times a track interruption occurs (compared to ground truth trajectory) due to miss-detection.
\end{itemize}
\noindent
True positives are detections which have not less than 50\% overlap when compared to their corresponding ground-truth bounding boxes. Each of these evaluation metrics have been explained well in~\cite{MOT16}.

\textbf{Implementation Details: } We implemented our proposed online tracking algorithm in MATLAB (not well optimized) on a i7 2.80 GHz core processor with 8 GB RAM, and it runs at about 3.3 fps. We trained our VerIdNet model on a NVIDIA GeForce GTX 1050 GPU for 200 epochs (1 epoch is one sweep over all the training data), after which it generally converges, using MatConvNet~\cite{VedLen15}. The mini-batch size is set to 20. We initialize the learning rate to $10^{-4}$ for the first 75 epochs, to $10^{-5}$ for the next 75 epochs and to $10^{-6}$ for the last 50 epochs. For feature extraction, we transferred the computation of the forward propagation to the NVIDIA GeForce GTX 1050 GPU for inference during online tracking. This forward propagation for extracting CNN features and the track extraction step are the two most computationally expensive parts of our tracking algorithm.

\textbf{Benchmark Evaluations: } We make comprehensive performance evaluations of our proposed online tracker, HISP-DAL, and compare it against state-of-the-art (SOTA) online as well as offline tracking algorithms such as MHT-DAM~\cite{KimLiCip15}, MHT-bLSTM~\cite{KimLiReh18}, IOU17~\cite{BocEisSik17}, DP-NMS~\cite{PirRamFow11}, SMOT~\cite{DicCamSzn13}, CEM~\cite{MilRotSch14}, JPDA-m~\cite{RezMilZha15}, EAMTT~\cite{MatPoiCav16}, GM-PHD-HDA~\cite{SonJeo16}, GMPHD-KCF~\cite{KutBosEis17}, GM-PHD~\cite{EisArpPat12}), GM-PHD-N1T~\cite{BaiWal19}, GM-PHD-DAL~\cite{Nat19}, HISP-T~\cite{Nat18}, JCmin-MOT~\cite{BorJeo17}, SAS-MOT17~\cite{MakFua19}, FPSN~\cite{LeeKim19}, OTCD-1~\cite{LiuWuLi19}, SORT17~\cite{BewGeOtt16}, AMIR~\cite{SadAli17}, KCF16~\cite{PenHen19} and OVBT~\cite{BanBaAla16}.

We evaluate our proposed online tracker quantitatively and compare it with other tracking algorithms in Table~\ref{tbl:MOT16} on MOT16 benchmark data set. As shown in this Table, HISP-DAL performs better than not only several online trackers listed in the table but also many offline trackers in terms of MOTA value. Though AMIR and KCF16 perform better than our tracker in terms of MOTA value, our tracker is much faster than them i.e. 3.3 times faster than AMIR and 33 times faster than KCF16. Our online tracker is ranked 2nd in terms of MOTP, and it also performs much better than several online and batch tracking algorithms such as GM-PHD-HDA, JPDF-m, DP-NMS, JCmin-MOT and GM-PHD-DAL when compared using MT, ML and FN values. Similarly, IDF1 of our tracker is higher than several online and offline tracking methods. As shown in the Table, the number of FP, IDSw and Frag is also much lower than numerous online and offline tracking approaches. Our tracker also gives promising results on MOT17 benchmark data set as is quantitatively shown in Table~\ref{tbl:MOT17}. It outperforms all other trackers in the table in MOTA. In terms of ML and FN, our tracker is ranked 2nd. Our tracker outperforms many online as well as offline trackers in both IDF1 and MOTP as well. Similarly, the number of FP, IDSw and Frag is also lower than many of the trackers in the table. As shown in the Table~\ref{tbl:MOT17}, our tracker which operates online outperforms the offline MHT-based trackers such as MHT-bLSTM in terms of FP and ML. Thus, our tracker has the advantage of not only working in an online fashion but also has better performance in some of the evaluation metrics than many of the offline trackers.

\begin{table*}
\begin{center}
\begin{tabular}{|l|c|c|c|c|c|c|c|c|c|c|r|}
\hline
Tracker & Tracking Mode & MOTA$\uparrow$ & MOTP$\uparrow$ & IDF1$\uparrow$ & MT (\%)$\uparrow$ & ML (\%)$\downarrow$ & FP$\downarrow$ & FN$\downarrow$ & IDSw$\downarrow$ & Frag$\downarrow$ &  Hz $\uparrow$  \\
\hline
MHT-DAM~\cite{KimLiCip15} & offline & \color{red}{\textbf{45.8}} & \color{blue}{\textbf{76.3}}	& \color{blue}{\textbf{46.1}} & \color{red}{\textbf{16.2}} &  \color{red}{\textbf{43.2}} &	6,412 & \color{red}{\textbf{91,758}}	& \color{blue}{\textbf{590}} & 781 & 0.8 \\
MHT-bLSTM6~\cite{KimLiReh18} & offline & \color{blue}{\textbf{42.1}} & 75.9 & \color{red}{\textbf{47.8}} & \color{blue}{\textbf{14.9}} & \color{blue}{\textbf{44.4}} & 11,637 & \color{blue}{\textbf{93,172}} & 753 & 1,156 & 1.8 \\
CEM~\cite{MilRotSch14} & offline & 33.2 & 75.8 & N/A & 7.8 &	54.4 &	6,837 &	114,322 & 642 &	\color{blue}{\textbf{731}} & 0.3 \\
DP-NMS~\cite{PirRamFow11} & offline & 32.2 & \color{red}{\textbf{76.4}} & 31.2 & 5.4 & 62.1 &	\color{red}{\textbf{1,123}} &	121,579 & 972 &	944 & \color{blue}{\textbf{5.9}} \\
SMOT~\cite{DicCamSzn13} & offline & 29.7 & 75.2 & N/A & 5.3 & 47.7 &	17,426 & 107,552 & 3,108 & 4,483 & 0.2 \\
JPDF-m~\cite{RezMilZha15} & offline & 26.2 & \color{blue}{\textbf{76.3}} & N/A & 4.1 &	67.5 &	\color{blue}{\textbf{3,689}} &	130,549 & \color{red}{\textbf{365}} & \color{red}{\textbf{638}} & \color{red}{\textbf{22.2}} \\
\hline
GM-PHD-HDA~\cite{SonJeo16} & \textbf{online} & 30.5 & 75.4 & 33.4 &	4.6 & 59.7 & 5,169 & 120,970 & \color{red}{\textbf{539}} & \color{red}{\textbf{731}} & \color{blue}{\textbf{13.6}} \\
AMIR~\cite{SadAli17}  & \textbf{online} &  \color{blue}{\textbf{47.2}} & 75.8	& \color{blue}{\textbf{46.3}} & \color{blue}{\textbf{10.6}} &	\color{blue}{\textbf{31.6}} & \color{blue}{\textbf{2,681}}	& \color{blue}{\textbf{92,856}} &	774 & 1,675 & 1.0 \\
OVBT~\cite{BanBaAla16} & \textbf{online} &  38.4 &	75.4 & 37.8	& 5.7 &	35.9 & 11,517 &	99,463	&	1,321 & 2,140 & 0.3 \\
KCF16~\cite{PenHen19} & \textbf{online} & \color{red}{\textbf{48.8}} &	75.7 & \color{red}{\textbf{47.2}} & \color{red}{\textbf{12.0}} & \color{red}{\textbf{28.9}} & 5,875 &	\color{red}{\textbf{86,567}} & 906 &	1,116 & 0.1 \\
HISP-T~\cite{Nat18} & \textbf{online} & 35.9 & 76.1 & 28.9 & 7.8 & 50.1 & 6,406 & 107,905 & 2,592 & 2,299 & 4.8 \\
GM-PHD-DAL~\cite{Nat19} & \textbf{online} & 35.1 & \color{red}{\textbf{76.6}} & 26.6 & 7.0 & 51.4 & \color{red}{\textbf{2,350}} & 111,886 & 4,047 & 5,338 & 3.5 \\
JCmin-MOT~\cite{BorJeo17} & \textbf{online} & 36.7 & 75.9 & 36.2 & 7.5 & 54.4 &  2,936	& 111,890 &	\color{blue}{\textbf{667}}	& \color{blue}{\textbf{831}} & \color{red}{\textbf{14.8}} \\
\textbf{HISP-DAL (ours)} & \textbf{online} & 37.4 & \color{blue}{\textbf{76.3}} & 30.5 & 7.6 & 50.9 & 3,222 & 108,865 & 2,101 & 2,151 & 3.3 \\
\hline
\end{tabular}
\end{center}
\caption{Tracking performance of representative trackers developed using both online and offline methods. All trackers are evaluated on the test data set of the \textbf{MOT16}~\cite{MOT16} benchmark using public detections. The first and second highest values are highlighted by $\color{red}{\textbf{red}}$ and $\color{blue}{\textbf{blue}}$, respectively (for both online and offline trackers). Evaluation measures with ($\uparrow$) show that higher is better, and with ($\downarrow$) denote lower is better. N/A shows not available.}
\label{tbl:MOT16}
\end{table*}

\begin{table*}[htbp]
\begin{center}
\begin{tabular}{|l|c|c|c|c|c|c|c|c|c|c|r|}
\hline
Tracker & Tracking Mode & MOTA$\uparrow$ & MOTP$\uparrow$ & IDF1$\uparrow$ & MT (\%)$\uparrow$ & ML (\%)$\downarrow$ & FP$\downarrow$ & FN$\downarrow$ & IDSw$\downarrow$ & Frag$\downarrow$ & Hz $\uparrow$\\
\hline
MHT-DAM~\cite{KimLiCip15} & offline & \color{red}{\textbf{50.7}} & \color{red}{\textbf{77.5}} & 47.2 & \color{red}{\textbf{20.8}} & \color{red}{\textbf{36.9}} & 22,875 & \color{red}{\textbf{252,889}} &	2,314 & \color{blue}{\textbf{2,865}} & 0.9\\
MHT-bLSTM~\cite{KimLiReh18} & offline & \color{blue}{\textbf{47.5}}	& \color{red}{\textbf{77.5}} & \color{blue}{\textbf{51.9}} & \color{blue}{\textbf{18.2}} &	41.7 &	25,981 & \color{blue}{\textbf{268,042}} & \color{blue}{\textbf{2,069}} & 3,124 & 1.9\\
IOU17~\cite{BocEisSik17} & offline & 45.5 & \color{blue}{\textbf{76.9}} & 39.4 & 15.7 & \color{blue}{\textbf{40.5}} & \color{blue}{\textbf{19,993}} & 281,643 & 5,988 & 7,404  & \color{red}{\textbf{1,522.9}} \\
SAS-MOT17~\cite{MakFua19} & offline & 44.2 & 76.4 & \color{red}{\textbf{57.2}} & 16.1 & 	44.3 &	29,473 & 283,611 & \color{red}{\textbf{1,529}} & \color{red}{\textbf{2,644}} & 4.8\\
DP-NMS~\cite{PirRamFow11} & offline & 43.7 & \color{blue}{\textbf{76.9}} & N/A & 12.6 & 46.5 & \color{red}{\textbf{10,048}} & 302,728 & 4,942 & 5,342 & \color{blue}{\textbf{137.7}} \\
\hline
FPSN~\cite{LeeKim19} & \textbf{online} & \color{blue}{\textbf{44.9}} & 76.6 & \color{red}{\textbf{48.4}} & \color{red}{\textbf{16.5}} & \color{red}{\textbf{35.8}} & 33,757 & \color{red}{\textbf{269,952}} & 7,136 & 14,491 & 12.0\\
EAMTT~\cite{MatPoiCav16} & \textbf{online} & 42.6 & 76.0 & 41.8 & 12.7 & 42.7 & 30,711 & 288,474 & 4,488 & 5,720 & 10.1\\
GMPHD-KCF~\cite{KutBosEis17} & \textbf{online} & 40.3 & 75.4 & 36.6 & 8.6 & 43.1 & 47,056 & 283,923 & 5,734 & 7,576 & 3.3\\
GM-PHD~\cite{EisArpPat12} & \textbf{online} & 36.2 & 76.1 & 33.9 & 4.2 & 56.6 & 23,682 & 328,526 & 8,025 & 11,972 & \color{blue}{\textbf{38.4}} \\
GMPHD-DAL~\cite{Nat19} & \textbf{online} & 44.4 & 77.4 & 36.2 & \color{blue}{\textbf{14.9}} & 39.4 & 19,170 & 283,380 & 11,137 & 13,900 & 3.4\\
OTCD-1~\cite{LiuWuLi19} & \textbf{online} & \color{blue}{\textbf{44.9}} & 77.4 & \color{blue}{\textbf{42.3}} & 14.0 &	44.2 &	\color{red}{\textbf{16,280}}	& 291,136 & \color{red}{\textbf{3,573}} & \color{blue}{\textbf{5,444}} & 5.5 \\
GMPHD-N1Tr~\cite{BaiWal19} & \textbf{online} & 42.1 & \color{blue}{\textbf{77.7}} & 33.9 &	11.9 &	42.7 &	\color{blue}{\textbf{18,214}}	& 297,646 &	10,698	& 10,864 & 9.9\\
SORT17~\cite{BewGeOtt16} & \textbf{online} & 43.1 &	\color{red}{\textbf{77.8}} & 39.8 & 12.5 &	42.3 &	28,398 & 287,582 &	4,852 &	7,127 & \color{red}{\textbf{143.3}} \\
GMPHD-SHA~\cite{SonJeo16} & \textbf{online} & 43.7	& 76.5 & 39.2 & 11.7 & 43.0 &	25,935 & 287,758 & \color{blue}{\textbf{3,838}} & \color{red}{\textbf{5,056}} & 9.2 \\
\textbf{HISP-DAL (ours)} & \textbf{online} & \color{red}{\textbf{45.4}} & 77.3 & 39.9 &	14.8 &	\color{blue}{\textbf{39.2}} &	21,820	& \color{blue}{\textbf{277,473}}	& 8,727 &	7,147 & 3.2 \\
\hline
\end{tabular}
\end{center}
\caption{Tracking performance of representative trackers developed using both online and offline methods. All trackers are evaluated on the test data set of the \textbf{MOT17} benchmark using public detections. The first and second highest values are highlighted by $\color{red}{\textbf{red}}$ and $\color{blue}{\textbf{blue}}$, respectively (for both online and offline trackers). Evaluation measures with ($\uparrow$) show that higher is better, and with ($\downarrow$) denote lower is better. N/A shows not available.}
\label{tbl:MOT17}
\end{table*}

\begin{figure*} [htb] 
  \begin{center}
  {\label{fig:DetectionFrame57} \includegraphics[height=0.272\textwidth]{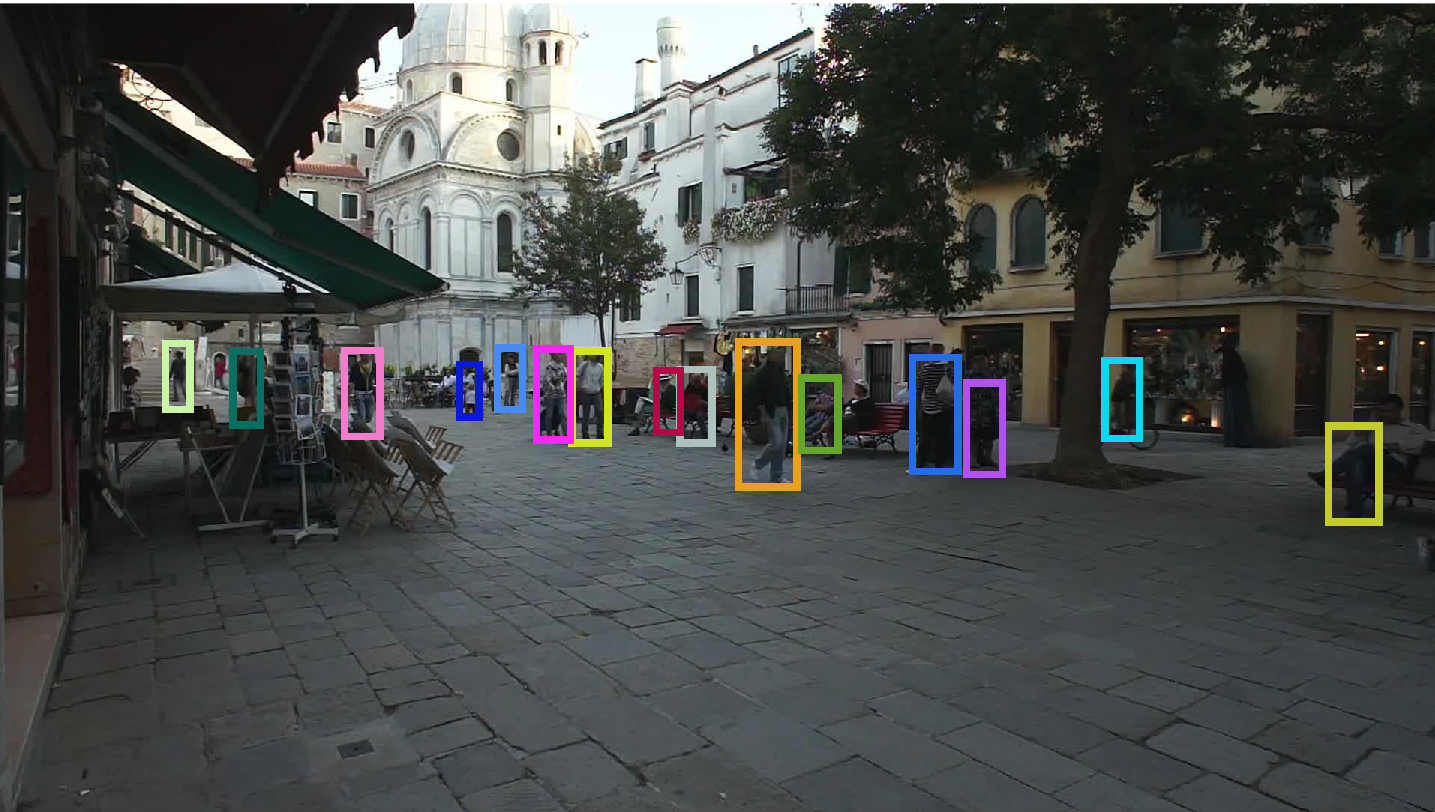}} 
  {\label{fig:TriPHDFrame57} \includegraphics[height=0.272\textwidth]{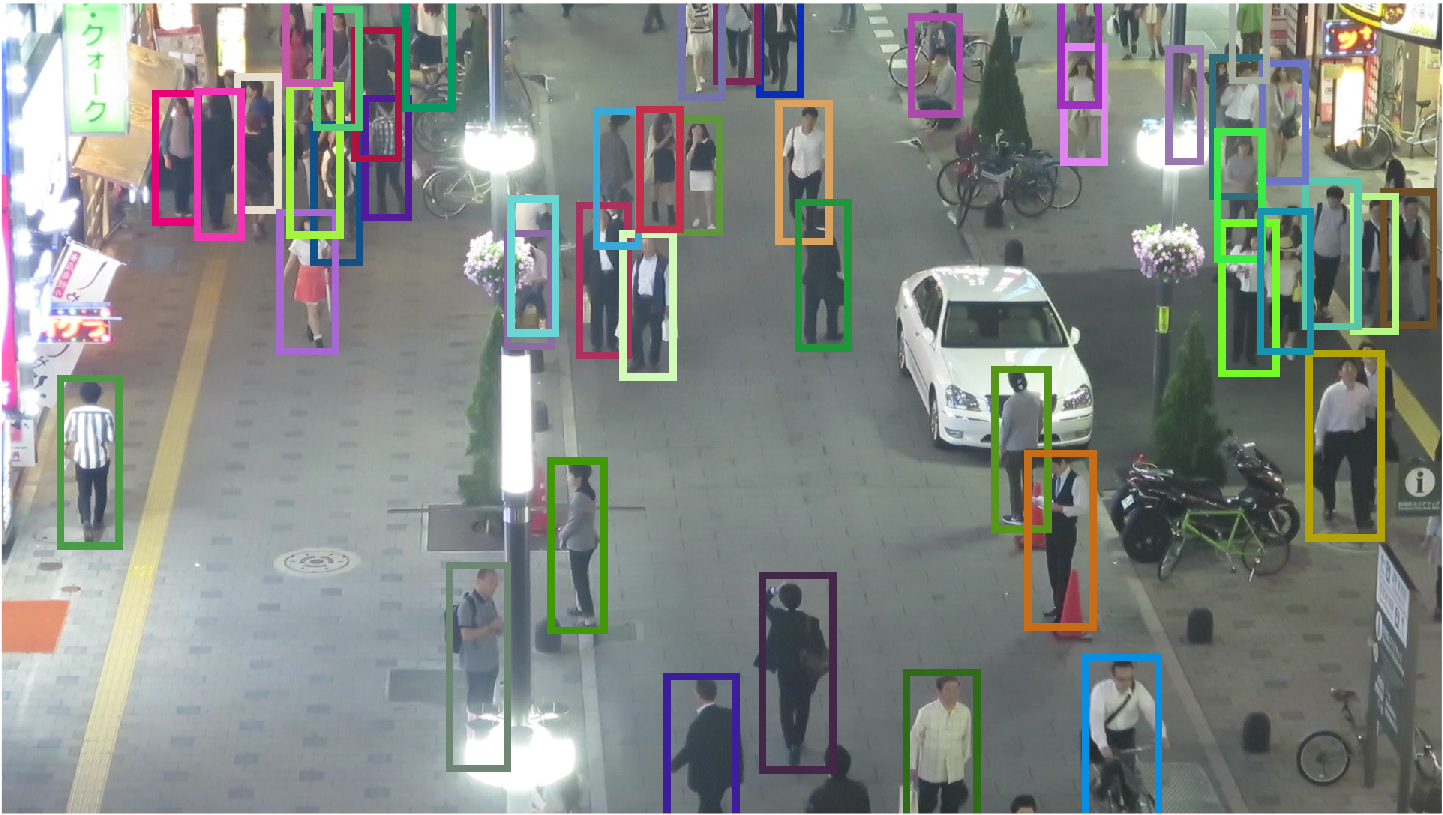}} \\
  {\label{fig:ThreePHDsFrame57} \includegraphics[height=0.31\textwidth]{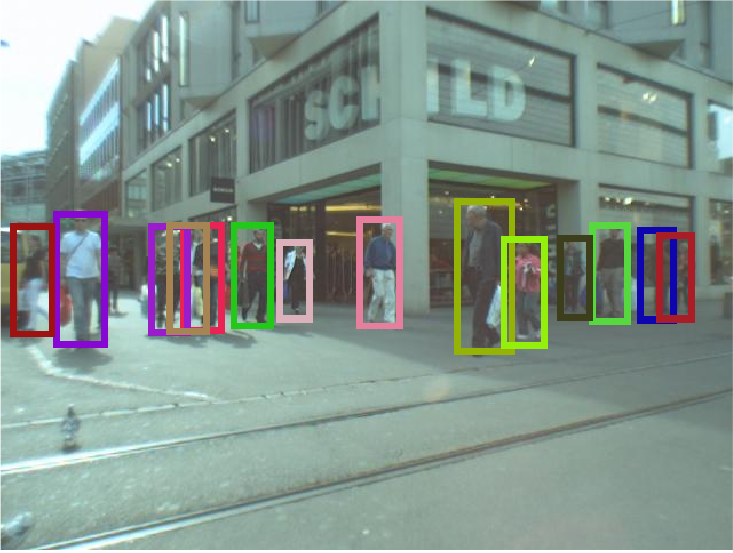}}  
  {\label{fig:ThreePHDsFrame57} \includegraphics[height=0.31\textwidth]{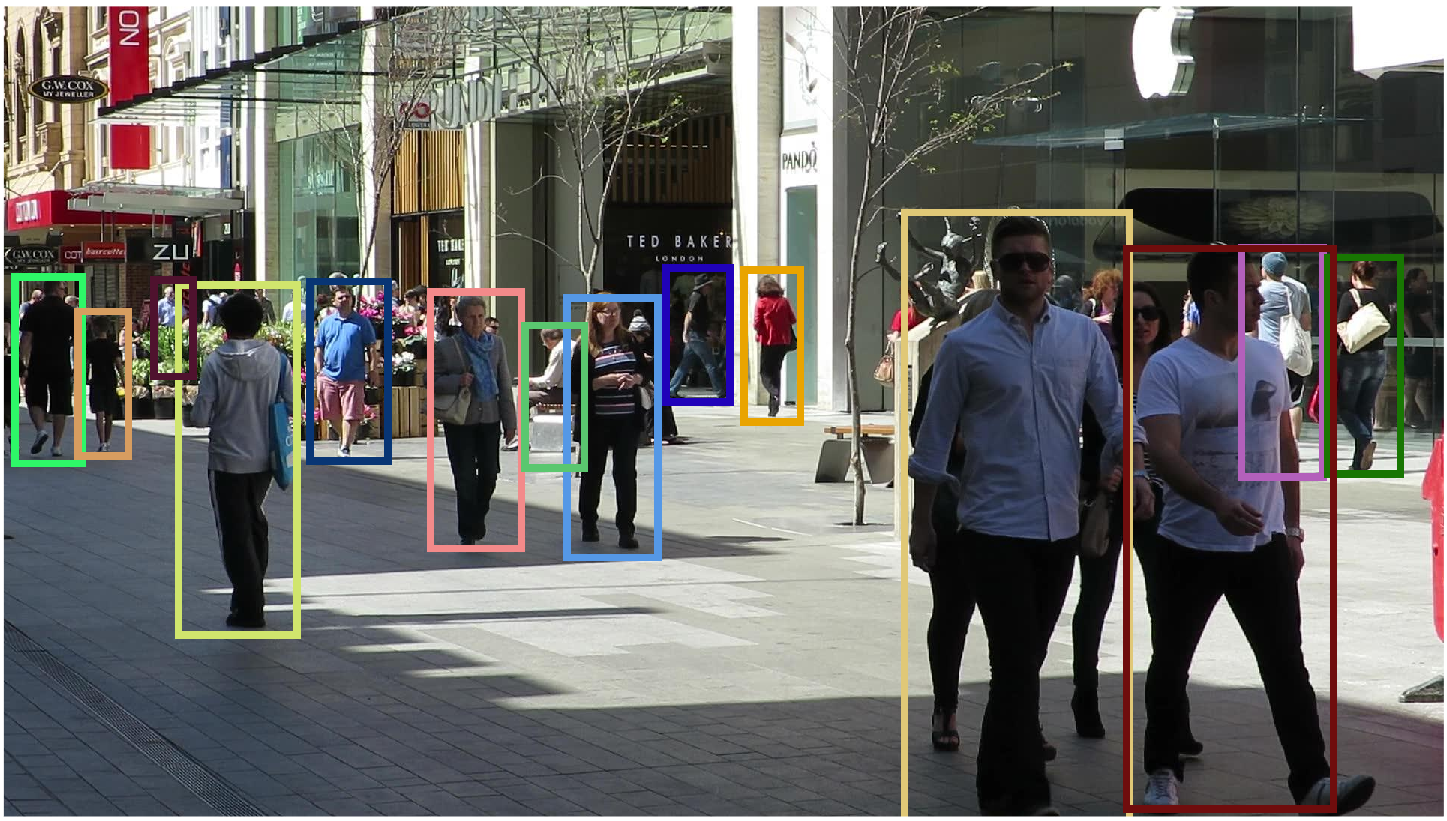}}\\ 
  {\label{fig:ThreePHDsFrame57} \includegraphics[height=0.27\textwidth]{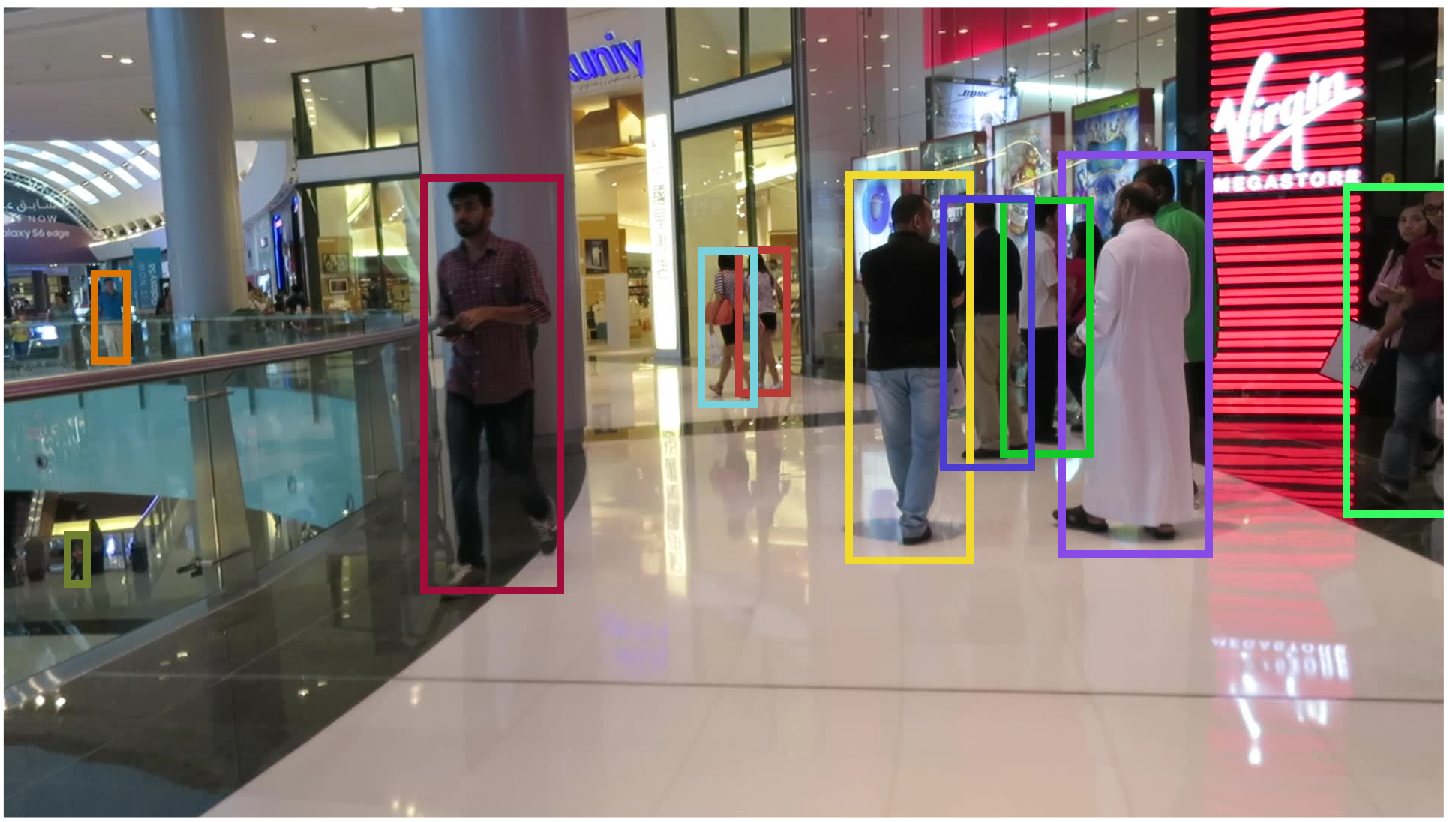}} 
  {\label{fig:ThreePHDsFrame57} \includegraphics[height=0.27\textwidth]{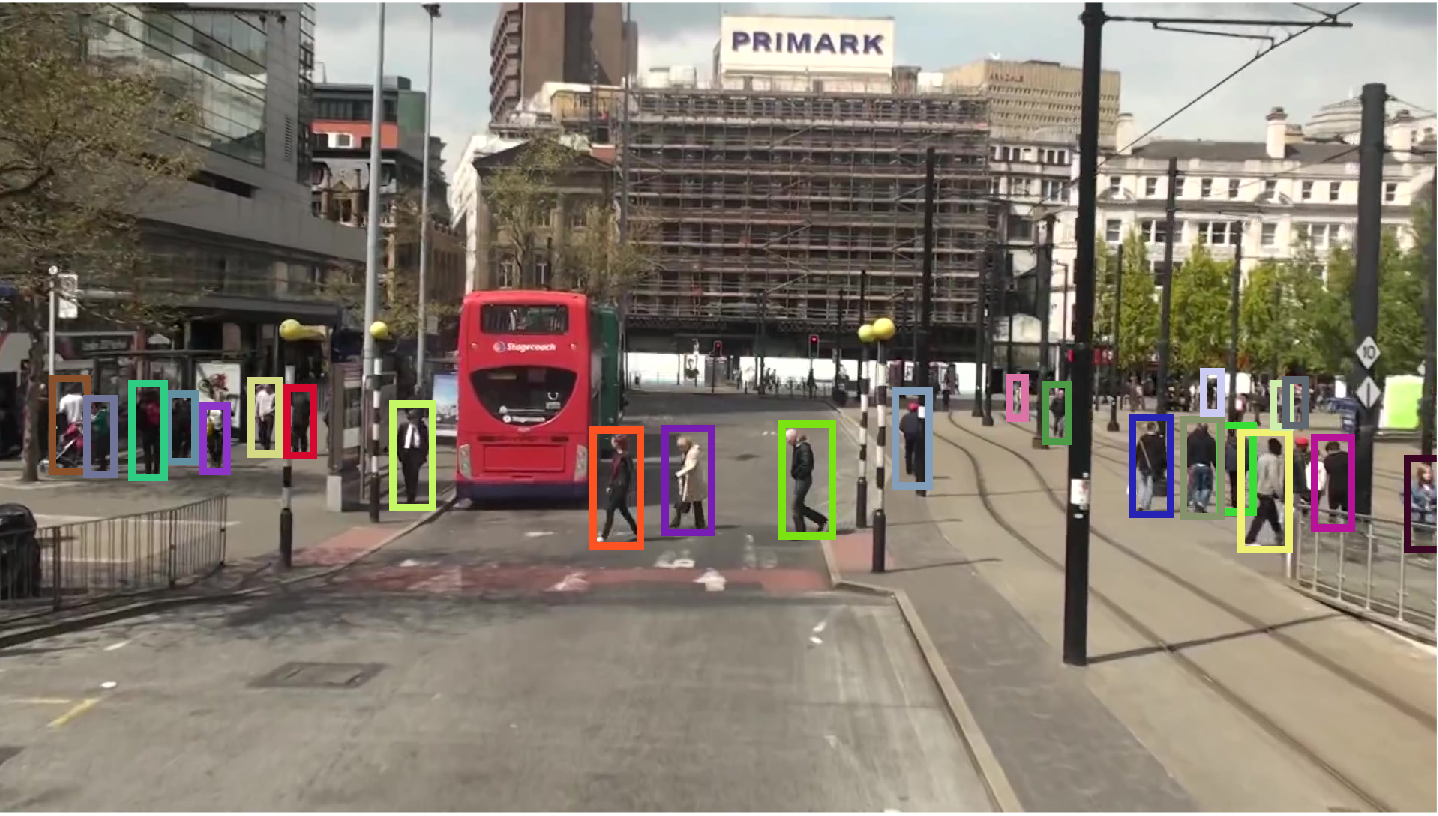}}  
  \end{center}
   \caption{\small{Sample results on several sequences of MOT17 data sets using SDP detector; bounding boxes
represent the tracking results with their color-coded identities. From left to right: MOT17-01-SDP and MOT17-03-SDP (top row), MOT17-06-SDP and MOT17-08-SDP (middle row), and MOT17-12-SDP and MOT17-14-SDP (bottom row). The videos of tracking results are available on the MOT Challenge website \color{red}{https://motchallenge.net/}.}}
  \label{fig:MOTA17}
\end{figure*}

\begin{figure*} [t]
  \begin{center}
  {\label{fig:DetectionFrame57} \includegraphics[width=0.620\linewidth]{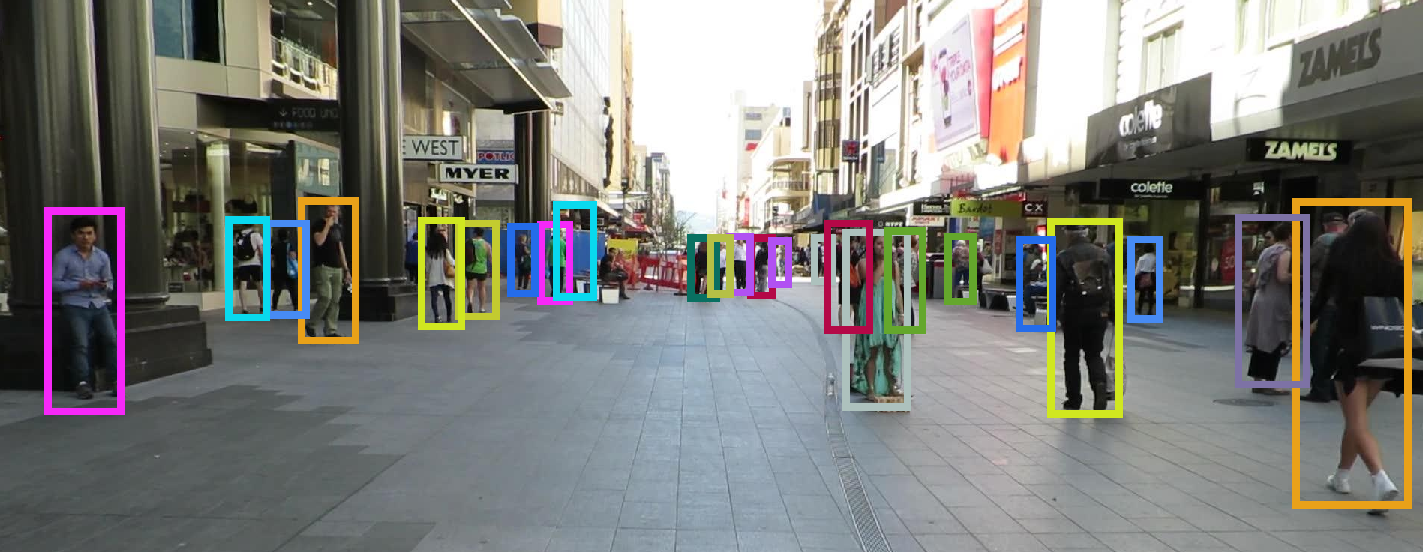}} \\
  {\label{fig:TriPHDFrame57} \includegraphics[width=0.620\linewidth]{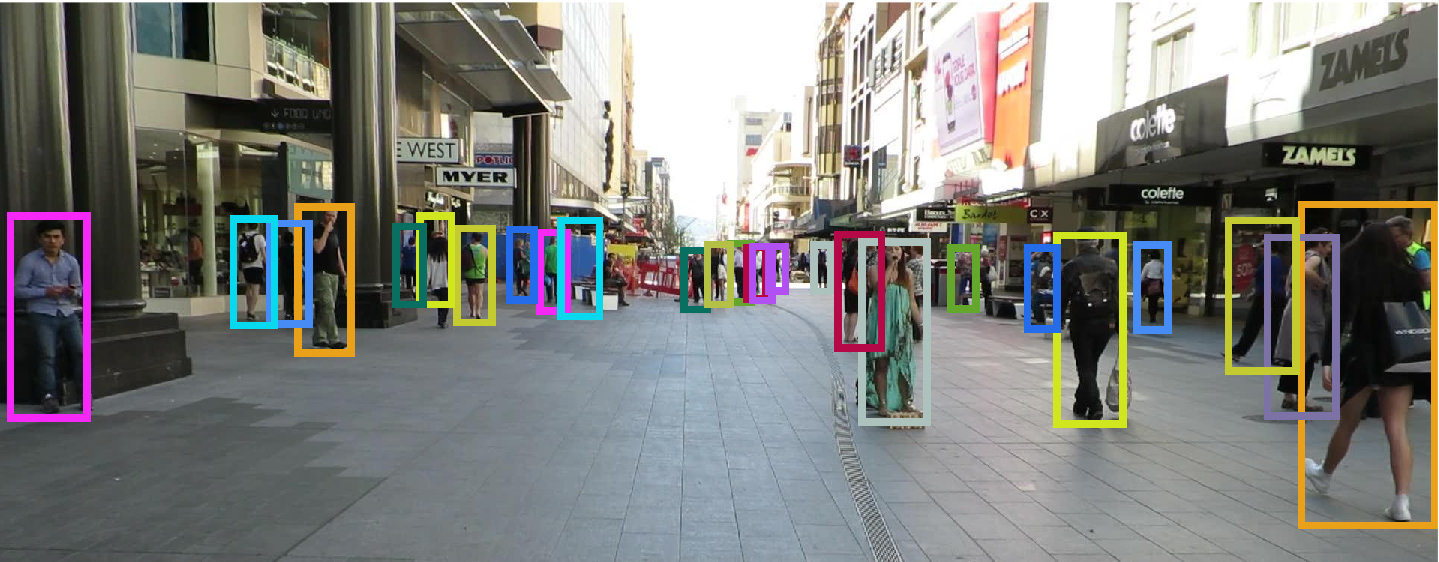}} \\ 
  {\label{fig:ThreePHDsFrame57} \includegraphics[width=0.620\linewidth]{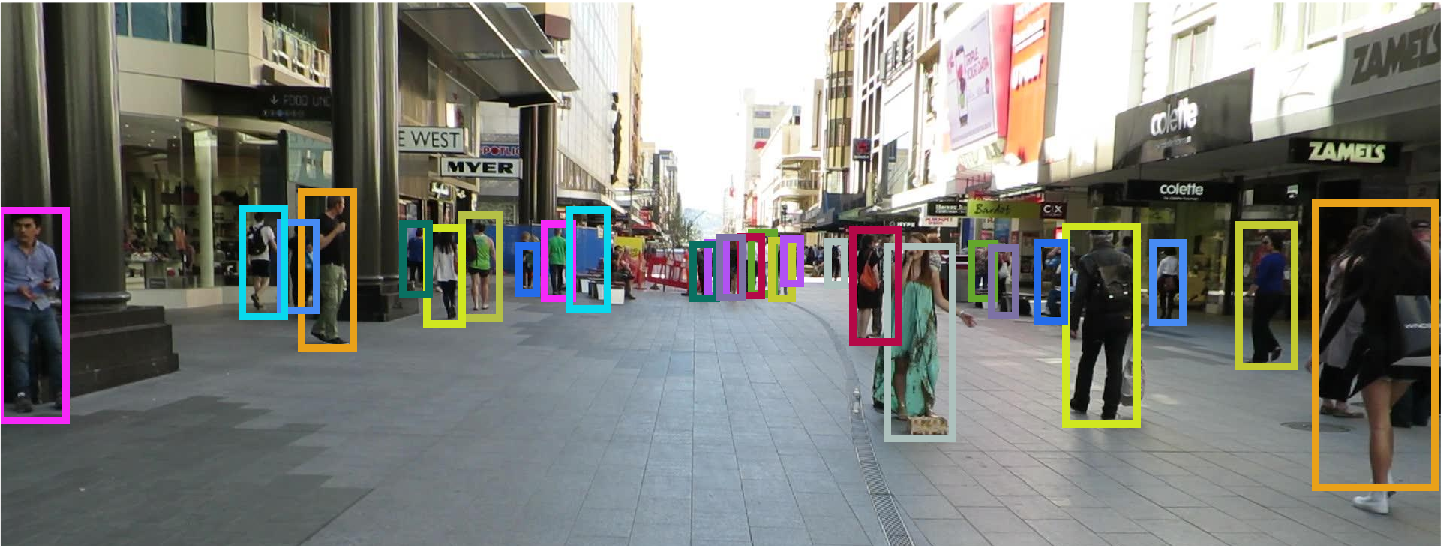}}  \\
  \end{center}
   \caption{\small{Sample results on the sequence MOT17-07-SDP (using SDP detector); bounding boxes represent the tracking results with their color-coded identities, for frames 376, 386 and 395 from top to bottom. The video of tracking results are available on the MOT Challenge website \color{red}{https://motchallenge.net/}.}}
  \label{fig:MOTA17-07-SDP}
\end{figure*}

We show some tracking results as examples using SDP detector and MOT17 test sequences in Figure~\ref{fig:MOTA17}. On the top row, MOT17-01-SDP and MOT17-03-SDP are shown from left to right. Similarly, in the middle row and on the bottom row are MOT17-06-SDP and MOT17-08-SDP, and MOT17-12-SDP and MOT17-14-SDP, respectively. We also show 3 frames of tracking results obtained from MOT17-07-SDP with our tracker in Figure~\ref{fig:MOTA17-07-SDP}. The identities of the tracking results in these figures are represented using colour-coded bounding boxes. Basically, a moving camera is used to capture the MOT17-07-SDP sequence which consists of 54 tracks from 500 frames. This sequence poses a challenge for a tracker because of a quite high density of pedestrians as well as significant motion of the camera which forces the trajectories of the pedestrians to be interrupted intermittently. Even on this sequence, our proposed online tracker has a reasonable performance with only very few identity switches because of significant motion of the camera and miss-detections.

\textbf{Ablation Study: } The incorporation of deep appearance information improves the tracking performance as shown in Table~\ref{tbl:MOT16}. Our proposed tracker, HISP-DAL, which uses the deep appearance features besides the spatio-temporal (motion) information outperforms the HISP-T tracker which uses only the motion information in terms of several evaluation metrics including MOTA, MOTP, IDSw, Frag and IDF1 (the HISP-T tracker is simply the HISP-DAL tracker without deep appearance information). For instance, our proposed tracker increases MOTA and IDF1 values from 35.9 and 28.9 to 37.4 and 30.5, respectively. This is an increase by 4.18\% for MOTA and by 5.54\% for IDF1. These increments in performance are obtained by incorporating the deeply learned appearance information into the HISP filter. Thus, the construction of an augmented likelihood in an elegant fashion using the deeply learned CNN appearance representation and spatio-temporal information improves the performance of the tracker in terms of many evaluation metrics.


\noindent


\noindent

\section{Conclusions} \label{Sec:Conclusion}

We have proposed a novel multi-object visual tracking algorithm using the recently introduced HISP filter. We apply this multi-target filter for tracking multiple objects in video frames captured under a range of environmental conditions and targets density. We employed a tracking-by-detection paradigm utilizing the public detections given in the MOT16 and MOT17 benchmark data sets. We also include a principled combination of motion and appearance information obtained from point detections and deep CNN features, respectively, into a single augmented likelihood for better tracking performance. Furthermore, we overcome the problem of two or more closely located objects sharing similar identity that occurs after the track extraction process in the HISP filter with the help of the weight of the confirmed hypothesis. Results demonstrate that our proposed tracker outperforms many of the state-of-the-art online and offline tracking algorithms on the MOT16 and MOT17 benchmark data sets with regards to many of the evaluation metrics such as tracking accuracy. This introduced online tracking algorithm can also be set as an offline tracking algorithm by applying the track extraction method to all the video frames rather than using a sliding window for better performance with its limitation for real-time applications. Using other optimization algorithms such as Lagrangian relaxation for the track extraction may also improve the performance. Making some changes to the network architecture such as replacing the softmax and cross entropy loss by the arcface~\cite{DenGuoXue19} and focal loss~\cite{TsuPriRos17}, respectively, may improve the discriminative learning of appearance features of targets, which we may look into in our future work. 


\section*{Acknowledgment}
We would like to thank Dr. Jeremie Houssineau for a brief discussion about the HISP filter.

\ifCLASSOPTIONcaptionsoff
  \newpage
\fi

\bibliographystyle{IEEEtran}
\bibliography{egbib}





\end{document}